\pdfoutput=1
\documentclass[letterpaper]{article}
\usepackage{rfbench}

\usepackage[hyphens]{url}
\usepackage{graphicx}
\usepackage{natbib}
\usepackage{caption}
\usepackage{booktabs}
\usepackage{amsmath}
\usepackage{array}
\usepackage{amssymb,bm,mathtools}
\usepackage{placeins}
\usepackage{fvextra}
\DeclareMathOperator{\diag}{diag}
\newcommand{\1}[1]{\mathbf{1}\!\left[#1\right]}

\title{RFBench: A Simulator-Grounded Multimodal Benchmark for Passive RF Layout Design and Understanding}

\author{
    Jiarui Cui\equalcontrib,
    Boxun Yan\equalcontrib
}
\affiliations{
    Department of Electrical and Computer Engineering, University of California, Los Angeles\\
    jiaruicui99@ucla.edu, boxun@ucla.edu
}

\begin{document}

\maketitle

\begin{abstract}
While multimodal language models can interpret diagrams and generate structured outputs, their capacity to connect passive radio-frequency (RF) design intent, layout geometry, and electromagnetic behavior remains largely untested. To address this gap, we introduce RFBench, a simulator-grounded dataset and benchmark for passive RF layout design and understanding. The dataset contains 10,000 two-port planar layouts simulated using ADS Momentum, with each layout represented by editable geometric primitives, a raster image, a $16\times16$ occupancy grid, complex S-parameters at 201 frequency points, a structural category, and an RF response label. Building on these aligned representations, we define seven tasks spanning specification-conditioned layout generation, representation translation, response prediction, simulator-based validation, design diagnosis and improvement, structural understanding, and response classification. Extensive baselines across seven multimodal language models reveal systematic weaknesses in generating structurally valid layouts and predicting electromagnetically faithful responses. We show that RF-specific fine-tuning increases T1a structural validity from 0.0\% to 83.2\% and the T3 primitive-input within-2-dB rate from 17.7\% to 62.8\%, demonstrating the value of domain adaptation while leaving considerable room for further progress. RFBench establishes a common testbed for evaluating multimodal RF capabilities and guiding the development of models that connect design intent and layout representations with physical behavior.
\end{abstract}

\begin{links}
    \link{Code}{https://github.com/jerry66s6/RFBench}
    \link{Dataset}{https://huggingface.co/datasets/Jerry1030/RFBench}
\end{links}

\section{Introduction}
Passive RF structures, including transmission lines, impedance-matching networks, passive filters, and antenna-tuning networks, are fundamental building blocks of high-frequency systems~\cite{hong2011microstrip}. Their electromagnetic (EM) behavior is strongly coupled to their geometry: changes in conductor dimensions, spacing, discontinuities, connectivity, and port placement can substantially alter frequency-dependent reflection and transmission. Designing such structures therefore requires repeated expert-guided iteration among performance specifications, editable metal layouts, rendered geometry, and full-wave electromagnetic simulation. This
process is computationally expensive and representation-fragmented, while a layout that appears visually plausible or syntactically valid may still be disconnected, improperly ported, or inconsistent with the desired response.

Machine learning (ML) has long supported surrogate modeling, inverse design, and optimization in RF and microwave CAD~\cite{zhang2003annrf,feng2022anncad, sedaghat2021compressed}, as part of broader machine learning for electronic design automation~\cite{rapp2022mlcad}. Recently, language-model methods and benchmarks attempted to address analog design, circuit reasoning, and hardware verification~\cite{lai2025analogcoder,wan2026fixme,skelic2025circuit, shi2025amsbench,li2025eee}. However, the field lacked a unified dataset and evaluation framework for testing whether multimodal models can reason across the representations central to passive RF design. Existing efforts primarily evaluated textual knowledge, schematic reasoning, code generation, or tool execution, rather than the relationship among RF specifications, planar geometry, and broadband electromagnetic response. Consequently, they cannot distinguish accurate RF reasoning from well-formatted or visually plausible outputs.

To address these gaps, we introduce \textbf{RFBench}, a simulator-grounded dataset and benchmark for multimodal passive RF layout design and understanding. Our main contributions are:

\begin{itemize}
    \item We construct a simulator-grounded RF-layout dataset of 10,000 two-port designs with aligned geometric primitives, raster images, occupancy grids, and complex $S$-parameter sweeps, addressing the lack of multimodal data connecting editable RF geometry with EM behavior.

    \item We build a seven-task multimodal benchmark with task-specific and ADS-grounded evaluation metrics spanning layout generation, representation translation, electromagnetic-response prediction, simulator validation, design diagnosis, structural understanding, and response classification.

    \item We conduct extensive experiments on seven general-purpose, open-weight, and circuit-specialized models to establish current baselines, identifying critical failure modes in geometric validity, cross-representation grounding, electromagnetic-response prediction, and specification-faithful layout generation.

    \item We implement RFBench-specific supervised fine-tuning and simulator-free reward shaping for Qwen3-VL-8B, leading to substantial improvements over the unadapted backbone in layout validity, representation translation, structural understanding, and EM prediction.
    
\end{itemize}

\begin{figure*}
    \centering
    \includegraphics[width=\textwidth]{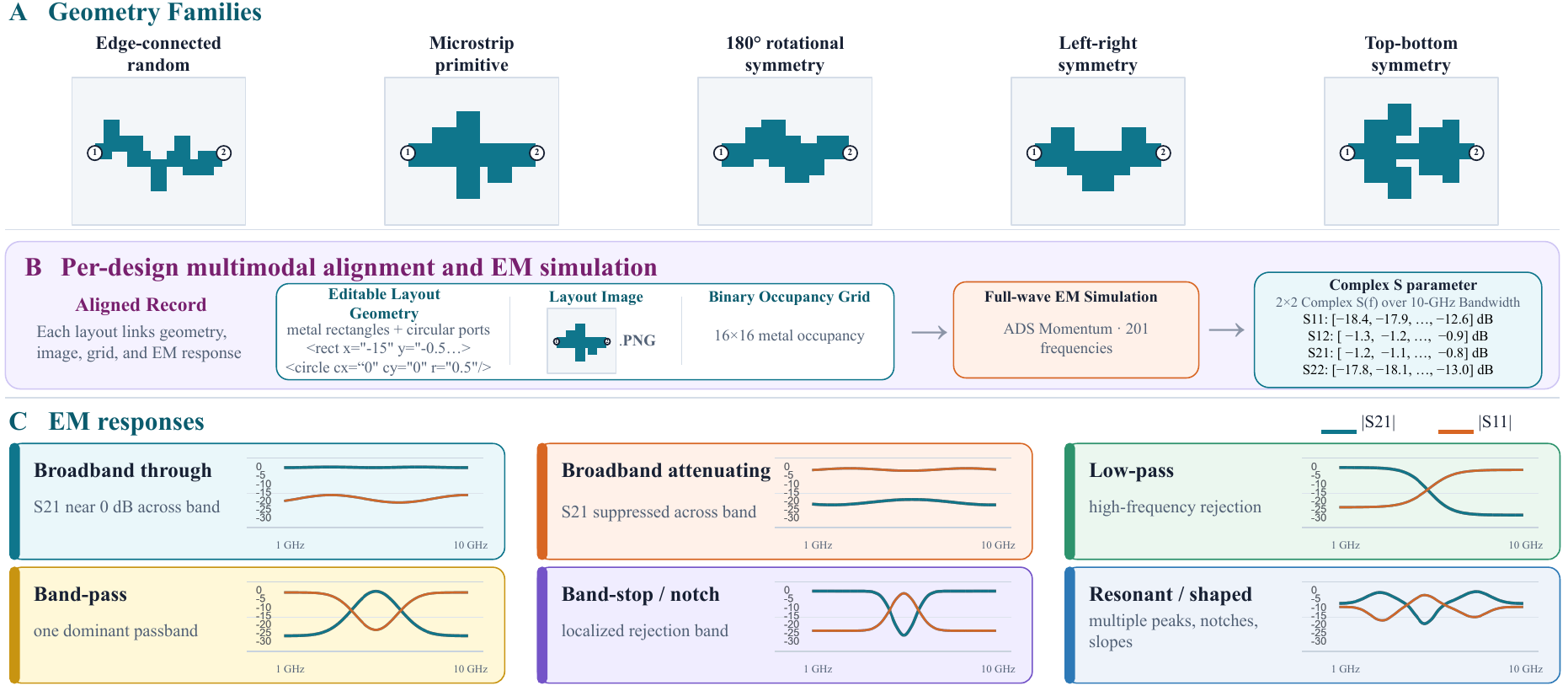}
    \caption{RFBench dataset construction and aligned multimodal record.
    (A) The corpus contains 10,000 two-port planar layouts distributed equally across five procedural structural categories.
    (B) For each design, a shared identifier aligns the editable
    geometric primitives, rendered image, $16\times16$ binary occupancy grid,
    and full complex $2\times2$ $S$-parameter sweep obtained from ADS Momentum
    at 201 frequency points.
    (C) Each simulated transmission response is assigned one of six deterministic RF behavior labels used for response classification. The displayed $S$-parameter curves are schematic class exemplars.}
    \label{fig:dataset_overview}
\end{figure*}

\section{Related Work}

\paragraph{Data-driven analog and RF design.}
Machine learning has supported RF computer-aided design through surrogate modeling, inverse design, and optimization~\cite{ zhang2003annrf,feng2022anncad}. These problems are challenging for high-frequency passive structures since conductor dimensions, discontinuities, spacing, and EM coupling could influence impedance and broadband scattering behavior. Neural operators provide a framework for learning parameterized PDE solution mappings~\cite{li2021fno}. NeurOLight applies this idea to frequency-domain photonic-device EM simulation~\cite{gu2022neurolight}, while recent physics-informed neural operators incorporate Maxwell-equation residuals during prediction~\cite{furat2025physicsinformed}. AICircuit provides simulator-derived data for analog/RF blocks~\cite{mehradfar2024aicircuit}, while CircuitNet and CircuitNet~3.0 provide tool-derived datasets for VLSI CAD~\cite{ chai2023circuitnet,wang2026circuitnet3}. Other methods map target responses to design parameters~\cite{sedaghat2021compressed} or combine method-of-moments simulation with tree search for pixelated diplexer design~\cite{lee2026pixelated}. These approaches show the value of simulation-grounded learning, but are formulated for specific device families, geometric parameterizations, or optimization objectives. They do not evaluate multimodal models across RF specifications, layout images, and broadband complex responses.

\paragraph{Language models for circuit design.} Language models broaden this setting by operating across circuit representations and engineering tools. AnalogCoder combines code generation with iterative simulation for analog synthesis~\cite{lai2025analogcoder}; ChatEDA connects an LLM to an RTL-to-GDSII flow, and ChipNeMo studies domain adaptation for chip-design assistance~\cite{he2023chateda,liu2023chipnemo}. CIRCUIT evaluates circuit reasoning~\cite{skelic2025circuit}, AMSBench covers multimodal analog and mixed-signal tasks~\cite{shi2025amsbench}, and AnalogSeeker specializes an open model for analog knowledge and design~\cite{chen2025analogseeker}. Analog Design Bench evaluates coding agents on 50 expert-contributed transistor-level tasks, in which agents edit and size SPICE netlists through multi-hour ngspice loops and an isolated verifier checks specification-based electrical tests~\cite{analogdesignbench2026}. Its tasks are defined at the schematic level, with layout and parasitics outside scope; RFBench instead targets passive planar layouts, whose broadband complex $S$-parameters are set by metal geometry and EM coupling rather than device sizing, and pairs evaluation with a 9,000-design training split for domain adaptation. These efforts primarily assess textual knowledge, schematic reasoning, code or netlist generation, and tool-flow execution. They do not test whether a model can preserve geometry across symbolic and visual representations, recognize RF layout constraints, and infer dense frequency-domain behavior.

\paragraph{Engineering benchmarks and multimodality.}
Machine-learning-for-EDA research emphasizes task-specific datasets, reliable metrics, and generalization across design settings~\cite{rapp2022mlcad}. MMMU evaluates expert multimodal reasoning across disciplines~\cite{ yue2024mmmu}, while EEE-Bench focuses on electrical and electronics problems and shows that models may overlook essential visual evidence~\cite{li2025eee}. Executable benchmarks such as VerilogEval and RTLLM evaluate generated RTL through simulation or testbenches~\cite{liu2023verilogeval, lu2023rtllmopensourcebenchmarkdesign}, and FIXME extends executable evaluation to hardware verification~\cite{wan2026fixme}. However, these benchmarks do not capture RF-specific challenges such as continuous planar geometry, feed-to-feed connectivity, geometric symmetry, and the dependence of broadband complex $S$-parameters on metal layout. They also do not distinguish symbolic or visual grounding failures from errors in EM response prediction.

\section{RFBench}

\subsection{RF Layout Dataset Construction}

\paragraph{Dataset composition and simulation.}
RFBench contains 10,000 two-port planar passive layouts, balanced across five structural categories with 2,000 designs each: \emph{edge-connected random}, \emph{microstrip-inspired primitives}, \emph{$180^\circ$ rotational symmetry}, \emph{left--right mirror symmetry}, and \emph{top--bottom mirror symmetry} (Figure~\ref{fig:dataset_overview}A). These categories provide a controlled design space spanning stochastic connectivity, transmission-line motifs, and measurable geometric symmetries of passive RF structures. Each design aligns an editable layout geometry of metal rectangles and circular ports with a layout PNG format image, a $16\times16$ binary occupancy grid, and its simulated EM response (Figure~\ref{fig:dataset_overview}B). Each layout is simulated in ADS Momentum at 201 equally spaced frequencies from 1 to 10~GHz, retaining the full complex $2\times2$ scattering matrix. Incomplete layout or responses are discarded, yielding 10,000 successfully simulated designs. The resulting transmission responses are assigned deterministic RF behavior labels, summarized in Figure~\ref{fig:dataset_overview}C.

\paragraph{Automated quality control.}
We apply deterministic consistency checks to every retained design. The final corpus contains 10,000 unique design identifiers, and 10,000 distinct primitive geometries, with no exact duplicates. It uses a common frequency grid, and contains finite $S$-parameters at every frequency point. Recomputing magnitude from the stored real and imaginary components differs from the retained magnitude by at most $5.6\times10^{-6}$~dB. We additionally verify that each occupancy grid exactly matches the rasterization of its corresponding geometry. These checks validate cross-representation alignment and numerical consistency.

\paragraph{Data splits.}
We construct a design-disjoint 90/10 train--test split using a random seed. The test set contains 1,000 layouts, balanced at 200 designs per structural category, while the remaining 9,000 designs form the training pool. No design identifier appears in both partitions. All common-set configurations for T1, T2, T3, T6, and T7 use the same 1,000 held-out designs across models. Task-specific sampling for T4 and T5 is described below. Full-corpus distributions and task-volume statistics are provided in Supplementary Figure~S1 and Table~S3. The RFBench dataset and code are publicly released under the CC BY 4.0 and MIT licenses, respectively.

\subsection{Benchmark Tasks}
\begin{figure*}[t]
    \centering
    \includegraphics[width=\textwidth]{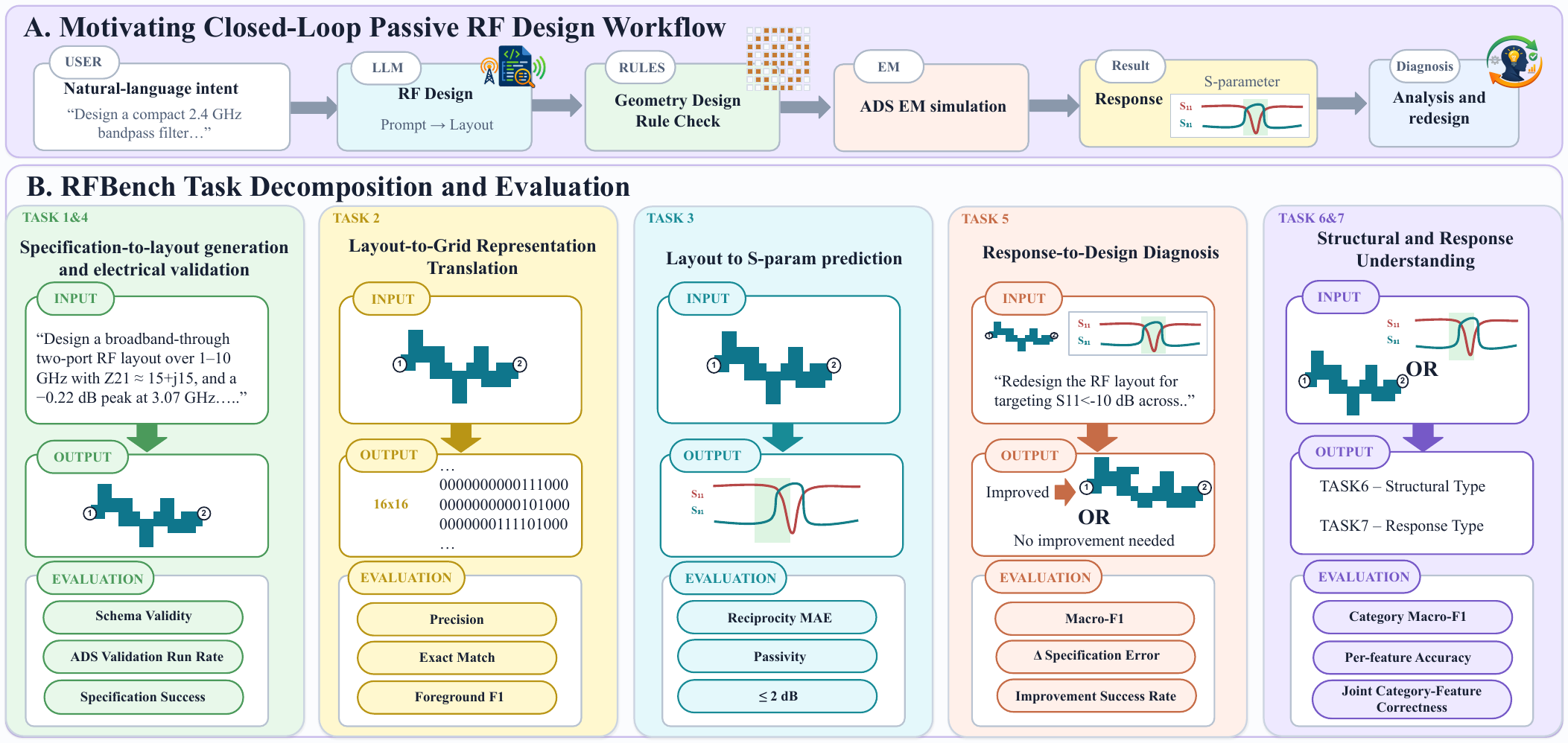}
    \caption{Motivating passive RF design workflow and RFBench task decomposition. (A) A motivating closed-loop workflow progresses from a natural-language RF specification to layout generation, structural validation, ADS Momentum simulation, and response-guided diagnosis and redesign. (B) RFBench decomposes this workflow into seven independently evaluated tasks, organized into five task groups. Each group shows its representative inputs, outputs, and evaluation dimensions, covering layout generation and simulator validation (T1 and T4), representation translation (T2), electromagnetic-response prediction (T3), design diagnosis and improvement (T5), and structural and response understanding (T6 and T7). Formal task definitions and evaluation metrics are provided in Table~\ref{tab:tasks}.}
    \label{fig:overview}
\end{figure*}

Building on the aligned dataset representations in Figure~\ref{fig:dataset_overview}, RFBench is motivated by the closed-loop passive RF design workflow illustrated in Figure~\ref{fig:overview}A, which progresses from a natural-language (NL) RF specification to layout generation, structural validation, electromagnetic simulation, and response-guided diagnosis and redesign. RFBench decomposes this workflow into seven independently evaluated tasks, organized into five task groups in Figure~\ref{fig:overview}B. Table~\ref{tab:tasks} summarizes the formal inputs, outputs, primary metrics, and evaluation sizes. T1a conditions generation on an RF specification, whereas T1b additionally specifies a structural category; for T2, T3, and T6, suffixes ``a'' and ``b'' denote primitive-geometry and image inputs, respectively.

\begin{table*}[t]
\centering
\small
\setlength{\tabcolsep}{1.2pt}
\begin{tabular}{@{}lllllr@{}}
\toprule
ID & Ability & Input & Output & Primary metric & Test $n$ \\
\midrule
T1a & Layout generation & RF response specification & Primitive layout & Structural validity & 1,000 \\
T1b &  Category-constrained generation & RF specification + category & Primitive layout & Structural validity; compliance & 1,000 \\
T2a & Representation translation & Primitive layout & $16\!\times\!16$ binary token & Foreground F1 & 1,000 \\
T2b & Representation translation & Layout image & $16\!\times\!16$ binary token & Foreground F1 & 1,000 \\
T3a & Response prediction & Primitive layout & Complex $2\!\times\!2$ $S$-parameter sweep & Complex/physics metrics & 1,000 \\
T3b & Response prediction & Layout image & Complex $2\!\times\!2$ $S$-parameter sweep & Complex/physics metrics & 1,000 \\
T4 & Simulator execution & T1a-generated layout & Run/fail + simulated response & Run rate; spec satisfaction & 250 \\
T5 & Improvement diagnosis & Layout + response + target & Decision + proposed redesign & Macro-F1/ADS improvement & 100 \\
T6a & Structural understanding & Primitive layout & Category + 10 feature flags & Accuracy/macro-F1 & 1,000 \\
T6b & Structural understanding & Layout image & Category + 10 feature flags & Accuracy/macro-F1 & 1,000 \\
T7 & Response classification & Complex $S$-parameter sweep & One of six response types & Accuracy/macro-F1 & 1,000 \\
\bottomrule
\end{tabular}
\caption{RFBench task taxonomy and primary comparison size.}
\label{tab:tasks}
\end{table*}

All common-set configurations use the same 1,000 held-out design IDs to enable paired comparisons across tasks and models; full-corpus 10,000 design results are reported in the supplementary material. T4 evaluates a fixed, category-balanced subset of 250 T1a targets. T5 uses the same 100 diagnosis cases for every model. All tasks are presented through NL instructions, with layout images accompanying textual inputs when applicable. The complete prompts, output schemas, and task availability are provided in the supplementary material.

\paragraph{T1: specification-conditioned layout generation.} Inverse RF design requires translating desired electrical behavior into physical geometry, a central problem in RF design and learned design optimization~\cite{feng2022anncad,sedaghat2021compressed, lee2026pixelated}. Each T1a prompt verbalizes the target magnitudes of $S_{11}$, $S_{12}$, $S_{21}$, and $S_{22}$ at three frequency landmarks, yielding 12 target values, and asks the model to generate an editable two-port layout. These targets are derived from an ADS-simulated design, so at least one realizing geometry exists. The reference geometry is withheld and exact reproduction is not required. T1b additionally specifies one of the five structural categories.

Both settings are evaluated for exact two-port count ($P$), canvas containment ($C$), and feed-to-feed connectivity ($K$), with structural validity defined as $V=P\wedge C\wedge K$. T1b further measures compliance with the requested category constraints.

\paragraph{T2: representation translation.}
Electronic design workflows move among symbolic, visual, and discrete representations. Prior work has explored multimodal alignment across circuit representations~\cite{fang2025circuitfusion}, while engineering and geometric-perception benchmarks reveal substantial limitations in models' use of visual evidence and spatial relationships~\cite{li2025eee,xing2024gepbench}. T2 evaluates this cross-representation grounding in passive RF layouts. Given an editable geometric description, T2a generates the corresponding $16\times16$ binary metal-occupancy grid; given a rendered layout image, T2b generates the same grid representation. Performance is evaluated over all 256 cells using foreground precision, recall, F1, and intersection-over-union. We additionally report normalized Hamming distance, exact-grid-match accuracy, and normalized metal-area difference. Comparing T2a and T2b distinguishes symbolic-serialization errors from visual-perception errors under a shared output representation.

\paragraph{T3: electromagnetic-response prediction.}
At RF and microwave frequencies, conductor dimensions, discontinuities, spacing, and electromagnetic coupling determine broadband reflection and transmission behavior~\cite{pozar2012microwave,hong2011microstrip}. Predicting this behavior from geometry is therefore a core objective of RF surrogate modeling and computer-aided design~\cite{zhang2003annrf,feng2022anncad}. Given either editable geometry (T3a) or a rendered image (T3b), the model predicts the full complex $2\times2$ scattering matrix at 201 frequencies. The main comparison reports complex MAE, the fraction of matrix--frequency points within 2~dB, passivity-valid frequency rate, complex reciprocity MAE, and spectral-derivative MAE relative to the ADS response. Full definitions and normalization checks are in the supplementary section ``T3 Metric Definitions and Official-Panel Diagnostics,'' with detailed component, magnitude, per-$S_{ij}$, and threshold metrics in Tables~S1 and S2.

\paragraph{Task-specialized surrogate reference.}
As an auxiliary reference, we evaluate the frozen CNN used for GRPO reward shaping on the same 1,000 held-out designs. The surrogate predicts the 201-point complex $S$-parameter sweep from layout geometry. It was trained on 8,100 non-test designs, with checkpoint selection on 900 disjoint validation designs. Unlike the instruction-following MLLMs, the surrogate is optimized solely for response regression. Its results therefore serve as a task-specialized reference for the learnability of the geometry-to-response relationship.

\paragraph{T4: closed-loop simulator validation.}
Executable evaluation is important because syntactically valid hardware outputs may still fail in the target tool flow or violate functional requirements~\cite{liu2023verilogeval,lu2023rtllmopensourcebenchmarkdesign, lai2025analogcoder,wan2026fixme}. T4 applies this principle to passive RF design by importing each T1a-generated layout into ADS Momentum and recording whether it produces a complete 201-frequency-point $S$-parameter sweep. Failed or missing simulations are counted against the full 250-design denominator. For successful simulations, performance is evaluated by comparing the predicted and target magnitudes of $S_{11}$, $S_{12}$, $S_{21}$, and $S_{22}$
at the three frequency landmarks specified in T1a. We report per-parameter and aggregate landmark MAE, together with a strict specification-success rate that requires all 12 values to lie within 3~dB and treats simulator failures as unsuccessful. T4 therefore distinguishes simulator executability from realization of the requested EM response. 

\paragraph{T5: design diagnosis and improvement.}
T5 provides a layout, its simulated response, and a target specification; the model determines whether improvement is needed and, when appropriate, proposes a revised layout. We evaluate the binary decision using accuracy, positive-class precision and recall, and macro-F1, and independently simulate proposed redesigns in ADS Momentum. Defining the target-error reduction as \(\Delta E=E_{\mathrm{original}}-E_{\mathrm{redesign}}\), we report ADS run rate, conditional improvement rate, median \(\Delta E\), and end-to-end ADS improvement over all ground-truth positive cases, counting missing, invalid, and simulation-failed redesigns as end-to-end failures.

\paragraph{T6: structural understanding.}
Reliable interpretation of layout structure supports constraint checking, representation alignment, and subsequent design operations. More generally, multimodal benchmarks have shown that models often struggle with geometric shapes, spatial relationships, and information embedded in engineering figures~\cite{li2025eee,shi2025amsbench,xing2024gepbench}. Given either
editable geometry (T6a) or a rendered image (T6b), the model predicts one of the five structural categories together with feature flags describing connectivity, symmetry, and geometric motifs. The performance is evaluated using category accuracy and macro-F1, per-feature accuracy, mean accuracy over the nine primary non-diagonal flags, exact feature-vector match, and joint category-plus-feature correctness.

\paragraph{T7: RF-response classification.}
Passive RF networks are commonly interpreted through qualitative transmission behaviors, such as low-pass, band-pass, and band-stop responses, which summarize how energy is transmitted across frequency~\cite{pozar2012microwave, hong2011microstrip}. T7 provides the simulated complex $S$-parameter sweep, and the model classifies it into one of six deterministic response types: \emph{broadband-through}, \emph{broadband-attenuating}, \emph{low-pass}, \emph{band-pass}, \emph{band-stop/notch}, or \emph{resonant-or-shaped}. We report six-way accuracy and macro-F1, emphasizing macro-F1 because the response classes are imbalanced.

Because the task families have heterogeneous outputs, metrics, and physical interpretations, RFBench does not treat a single aggregate score as the primary leaderboard criterion. Exploratory cross-task aggregations and sensitivity analyses are reported in Supplementary Tables~S13 and S14.

\begin{table*}[t]
\centering
\small
\setlength{\tabcolsep}{1.8pt}
\begin{tabular*}{\textwidth}{@{\extracolsep{\fill}}lrrrrrrrrrrr@{}}
\toprule
& \multicolumn{3}{c}{T1a: generation} & \multicolumn{4}{c}{T1b: constrained generation} & \multicolumn{4}{c}{T4: ADS validation} \\
\cmidrule(lr){2-4}\cmidrule(lr){5-8}\cmidrule(lr){9-12}
Model & Ports $\uparrow$ & On metal $\uparrow$ & Valid $\uparrow$ & Ports $\uparrow$ & On metal $\uparrow$ & Valid $\uparrow$ & Full spec. $\uparrow$ & Run $\uparrow$ & $S_{11}$ MAE $\downarrow$ & $S_{22}$ MAE $\downarrow$ & Strict $\uparrow$ \\
\midrule
AnalogSeeker & 98.5 & 77.7 & 0.0 & 88.8 & 40.2 & 0.0 & 15.1 & 77.2 & 8.94 & 9.00 & 0.0 \\
Claude 4.5 & 100.0 & 88.2 & 0.0 & 100.0 & 77.7 & 0.0 & 38.9 & 87.6 & 8.34 & 8.34 & 0.4 \\
GPT-5.4-mini & 99.9 & 60.0 & 0.0 & 100.0 & 47.8 & 0.0 & 0.0 & 61.6 & 7.39 & 7.38 & 0.8 \\
Llama 3.2-11B & 70.1 & 33.7 & 0.0 & 95.0 & 11.2 & 0.0 & 0.1 & 37.2 & 8.19 & 8.18 & 0.4 \\
Qwen3-VL-8B & 100.0 & 0.2 & 0.0 & 98.6 & 34.8 & 0.0 & 0.8 & 86.8 & 8.31 & 8.33 & 0.0 \\
\midrule
Qwen3 SFT & 100.0 & \textbf{100.0} & \underline{82.7} & 100.0 & \textbf{100.0} & \underline{86.0} & \textbf{85.7} & \textbf{100.0} & \underline{7.02} & \underline{7.05} & \underline{1.6} \\
Qwen3 GRPO & 100.0 & \textbf{100.0} & \textbf{83.2} & 100.0 & \textbf{100.0} & \textbf{86.4} & \underline{85.2} & \textbf{100.0} & \textbf{6.79} & \textbf{6.82} & \textbf{2.8} \\
\bottomrule
\end{tabular*}
\caption{Combined T1/T4 results. T1a and T1b use the shared 1,000-design panel; T4 uses its balanced 250-target subset. Entries are percentages except conditional $S$-parameter landmark MAE in dB. ``Full spec.'' requires all measurable category constraints; ``Strict'' requires all 12 T4 landmarks within 3~dB and counts ADS failures as unsuccessful.}
\label{tab:t1-t4}
\end{table*}

\begin{table*}[t]
\centering
\small
\setlength{\tabcolsep}{0.8pt}
\begin{tabular*}{\textwidth}{@{\extracolsep{\fill}}llrrrrrrrrrrrrr@{}}
\toprule
& & \multicolumn{4}{c}{T2: representation translation}
& \multicolumn{5}{c}{T3: response prediction}
& \multicolumn{4}{c}{T6: structural understanding} \\
\cmidrule(lr){3-6}\cmidrule(lr){7-11}\cmidrule(lr){12-15}
Model & Input
& Prec. $\uparrow$
& Rec. $\uparrow$
& F1 $\uparrow$
& Exact $\uparrow$
& C-MAE $\downarrow$
& $\leq2$ dB $\uparrow$
& Pass. $\uparrow$
& Recip. $\downarrow$
& $\Delta$-MAE $\downarrow$
& Cat. acc. $\uparrow$
& Cat. F1 $\uparrow$
& Feat. $\uparrow$
& Joint $\uparrow$ \\
\midrule
AnalogSeeker & primitive
& 29.2 & 9.4 & 14.2 & 0.0
& .901 & 13.4 & 98.6 & .0180 & .0437
& 21.6 & 10.9 & 55.7 & 7.4 \\

Claude 4.5 & primitive
& 26.9 & 37.6 & 31.4 & 0.0
& .873 & 28.0 & 30.9 & .0004 & .0575
& 51.1 & 46.7 & 61.7 & 18.6 \\

& image
& 30.4 & 56.1 & 39.4 & 0.0
& .873 & 27.0 & 32.5 & .0510 & .0706
& 48.0 & 46.7 & 66.7 & 12.3 \\

GPT-5.4-mini & primitive
& 36.1 & 23.9 & 28.7 & 0.0
& .796 & 28.4 & 79.4 & .2898 & .0452
& 22.6 & 13.0 & 55.2 & 15.3 \\

& image
& 48.4 & 47.9 & 48.2 & 0.0
& .714 & 11.6 & 90.0 & .1719 & .0483
& 25.9 & 16.1 & 58.1 & 19.3 \\

Llama 3.2-11B & primitive
& 20.1 & 20.4 & 20.3 & 0.0
& .638 & 0.2 & \textbf{100.0} & .0024 & .0432
& 18.8 & 12.3 & 72.2 & 16.0 \\

& image
& 19.1 & 7.3 & 10.5 & 0.0
& .667 & 3.1 & 97.6 & .0460 & .0521
& 19.9 & 7.5 & 56.5 & 13.6 \\

Qwen3-VL-8B & primitive
& 24.3 & 0.5 & 1.0 & 0.0
& .829 & 17.7 & 74.2 & .9712 & .0465
& 22.9 & 19.2 & 61.2 & 6.5 \\

& image
& 20.8 & 0.3 & 0.7 & 0.0
& .730 & 11.1 & \underline{98.5} & .8074 & .0446
& 36.7 & 25.1 & 69.9 & 36.7 \\

Qwen3 SFT & primitive
& \textbf{99.99} & \textbf{100.00} & \textbf{99.99} & \underline{99.4}
& .304 & 62.6 & 74.0 & $<10^{-6}$ & .0334
& \textbf{99.4} & \textbf{99.4} & \textbf{99.8} & \textbf{99.4} \\

& image
& 99.98 & 99.97 & 99.98 & 99.1
& \underline{.291} & \textbf{64.1} & 72.0 & $<10^{-6}$ & \underline{.0327}
& 98.8 & 98.8 & 99.7 & 98.8 \\

Qwen3 GRPO & primitive
& \textbf{99.99} & \textbf{100.00} & \textbf{99.99} & \textbf{99.5}
& .303 & 62.8 & 73.5 & $<10^{-6}$ & .0333
& \underline{99.3} & \underline{99.3} & \textbf{99.8} & \underline{99.3} \\

& image
& 99.97 & 99.97 & 99.97 & 99.0
& \textbf{.290} & \textbf{64.1} & 71.5 & $\mathbf{<10^{-7}}$ & \textbf{.0326}
& 99.0 & 99.0 & \textbf{99.8} & 99.0 \\

\midrule
Surrogate $^\dagger$ & layout
& -- & -- & -- & --
& .190 & 84.4 & 28.3 & .1159 & .0933
& -- & -- & -- & -- \\
\bottomrule
\end{tabular*}
\caption{Combined T2/T3/T6 results on the shared 1,000-design panel. Percentage columns include T2 and T6 scores, the fraction of all 804 T3 values within 2~dB, and passivity-valid frequencies. T3 C-MAE and spectral-derivative $\Delta$-MAE cover all four complex entries; Recip. is mean complex $\lvert S_{12}-S_{21}\rvert$. $^\dagger$The surrogate is task specialized.}
\label{tab:t2-t3-t6}
\end{table*}

\begin{table*}[t]
\centering
\small
\setlength{\tabcolsep}{3.0pt}
\begin{tabular*}{\textwidth}{@{\extracolsep{\fill}}lrrrrrrrrrr@{}}
\toprule
& \multicolumn{8}{c}{T5: diagnosis/redesign} & \multicolumn{2}{c}{T7: classification} \\
\cmidrule(lr){2-9}\cmidrule(lr){10-11}
Model & Acc. $\uparrow$ & Prec. $\uparrow$ & Rec. $\uparrow$ & Macro-F1 $\uparrow$ & \shortstack{Solved/\\candidates} & \shortstack{Improved/\\solved} & \shortstack{End-to-end ADS\\improv. (\%) $\uparrow$} & \shortstack{Median $\Delta E$\\(dB) $\uparrow$} & Acc. $\uparrow$ & Macro-F1 $\uparrow$ \\
\midrule
AnalogSeeker & \underline{85.0} & \underline{81.8} & 98.4 & \underline{82.0} & 63/63 & 46/63 (73.0\%) & \textbf{71.9} & \underline{+4.860} & 7.9 & 7.5 \\
Claude 4.5 & 73.0 & 70.3 & \textbf{100.0} & 61.3 & 63/64 & 45/63 (71.4\%) & \underline{70.3} & +4.445 & 47.8 & 36.4 \\
GPT-5.4-mini & \textbf{89.0} & \textbf{94.9} & 87.5 & \textbf{88.4} & 56/56 & \textbf{43/56 (76.8\%)} & 67.2 & +4.427 & 25.3 & 16.8 \\
Llama 3.2-11B & 64.0 & 64.0 & \textbf{100.0} & 39.0 & 58/64 & \underline{44/58 (75.9\%)} & 68.8 & +4.766 & 27.9 & 7.3 \\
Qwen3-VL-8B & 56.0 & 64.7 & 68.8 & 51.0 & 44/44 & 31/44 (70.5\%) & 48.4 & \textbf{+5.419} & 16.8 & 14.8 \\
Qwen3 SFT & 64.0 & 69.4 & 78.1 & 58.6 & 50/50 & 29/50 (58.0\%) & 45.3 & +0.007 & \underline{94.2} & \underline{74.0} \\
Qwen3 GRPO & 63.0 & 67.5 & 81.3 & 55.5 & 44/52 & 23/44 (52.3\%) & 35.9 & +0.044 & \textbf{94.3} & \textbf{74.2} \\
\bottomrule
\end{tabular*}

\caption{Combined T5 diagnosis, redesign, and T7 classification metrics. T5 contains 100 cases (64 positive and 36 negative), with ``needs improvement'' as the positive class. $\Delta E=E_{\mathrm{original}}-E_{\mathrm{redesign}}>0$. T7 contains 1,000 designs; classification columns are percentages and macro-F1 accounts for class imbalance.}
\label{tab:t5-t7-detail}
\end{table*}

\subsection{Baselines and Adaptation}

\paragraph{Baselines.}
We benchmark five representative models spanning access setting, multimodal capability, model scale, and circuit-domain specialization:
(i) \textbf{Closed General-Purpose Models}, including Claude Sonnet~4.5 and GPT-5.4-mini; (ii) \textbf{Open-Weight Multimodal Models}, including Llama~3.2-11B-Vision-Instruct and Qwen3-VL-8B-Instruct~\cite{qwen3vl}; and (iii) a \textbf{Domain-Specialized Circuit Model}, AnalogSeeker, an open-source Qwen2.5-32B-Instruct model adapted to analog-circuit knowledge~\cite{chen2025analogseeker}. Qwen3-VL-8B-Instruct also serves as the backbone for our SFT and GRPO variants, enabling direct comparison.

\paragraph{Supervised fine-tuning.}
We adapt Qwen3-VL-8B-Instruct using 4-bit QLoRA~\cite{dettmers2023qlora} for one epoch over 225,000 task conversations constructed from the 9,000-design training partition. Training minimizes assistant-token cross entropy while masking system and user tokens. Full optimization and adapter settings are provided in Supplementary Table~S11.

\paragraph{GRPO and RF-specific rewards.}
Starting from SFT, we apply GRPO~\cite{shao2024deepseekmath} with four outputs per prompt and task-specific rewards:
\begin{equation}
R(\tau,y)=
\begin{cases}
0.15v, & \tau=\mathrm{T1a},\ \text{invalid},\\
0.25v+0.75z, & \tau=\mathrm{T1a},\ \text{valid},\\
0.75F_{\mathrm{fg}}+0.25E_g, & \tau=\mathrm{T2a/b},\\
\operatorname{clip}(1-m/0.25,0,1), & \tau=\mathrm{T3}_{\mathrm{comp}},\\
\rho_5, & \tau=\mathrm{T5},\\
0.70c+0.30\bar f, & \tau=\mathrm{T6a/b},\\
q, & \tau\in\{\mathrm{T3}_{\mathrm{type}},\mathrm{T7}\},\\
0, & \text{otherwise}.
\end{cases}
\label{eq:grpo_reward}
\end{equation}
Here, $v$ measures geometric quality, $z$ measures specification satisfaction predicted by a frozen surrogate, and the remaining terms correspond to task-native correctness measures. The surrogate is trained and selected exclusively on disjoint non-test designs. It shapes T1a and T5 rewards only; all reported benchmark scores are computed from held-out model outputs or independent ADS evaluation. Detailed reward branches and launch commands are given in the supplementary sections ``Implemented GRPO Reward Function'' and ``Split Accounting and Adaptation Details''; Supplementary Table~S11 reports hyperparameters, and Supplementary Figure~S2 reports surrogate validation.

\section{Results and Discussion}

Tables~\ref{tab:t1-t4}--\ref{tab:t5-t7-detail} report results on the designated held-out evaluation sets. Following the workflow in Figure~\ref{fig:overview}, we organize the analysis around five transformations connecting RF design intent, layout representations, electromagnetic behavior, and design understanding.

\paragraph{Specification-to-layout generation and electrical validation (T1 and T4).}
Unadapted models frequently produce plausible-looking outputs but fail the combined structural requirements of layout generation. Although several models generate two port markers in nearly all T1a cases, every unadapted baseline obtains 0.0\% structural validity under the joint port, canvas, and connectivity checks. RFBench-specific adaptation changes this behavior: Qwen3 SFT reaches 82.7\% T1a validity and 86.0\% T1b validity, while Qwen3 GRPO reaches 83.2\% and 86.4\%, respectively. For category-conditioned generation, the adapted models achieve 85.7\% and 85.2\% full category compliance.

T4 shows that structural validity does not guarantee the requested electrical behavior. Both adapted models achieve a 100\% ADS run rate on 250 generated layouts, compared with 37.2--87.6\% for the unadapted baselines. However, strict specification success remains only 1.6\% for SFT and 2.8\% for GRPO when all 12 target values must fall within 3~dB. Thus, adaptation largely improves valid and simulatable layout generation, while inverse RF design remains challenging.

\paragraph{Layout-to-grid representation translation (T2).}
Unadapted models show limited ability to preserve layout geometry when translating either editable primitives or images into binary occupancy grids. The strongest unadapted foreground F1 scores are 31.4\% for primitive input and 48.2\% for image input, and no unadapted model produces an exact grid match. In contrast, Qwen3 SFT and GRPO achieve foreground F1 above 99.97\% for both modalities and exact-match rates between 99.0\% and 99.5\%. The small gap between primitive and image inputs after adaptation indicates that direct supervision can effectively align symbolic and visual layouts with a shared discrete representation. T2 therefore becomes nearly saturated after RFBench-specific training.

\paragraph{Layout-to-response prediction (T3).}
EM-response prediction remains more challenging than representation translation. Among unadapted models, Claude~4.5 is the most balanced across modalities, with 28.0\% and 27.0\% of matrix--frequency points within 2~dB for primitive and image inputs, while GPT-5.4-mini reaches 28.4\% and 11.6\%, respectively. Qwen3 SFT reduces complex MAE to .304 and .291, with within-2-dB rates of 62.6\% and 64.1\%. GRPO yields only marginal changes, reaching .303 and .290 complex MAE and 62.8\% and 64.1\% within 2~dB. Both adapted models also achieve near-zero reciprocity error and lower spectral-derivative error than the backbone. Passivity, however, must be interpreted jointly with response fidelity: unadapted Qwen3 on images attains 98.5\% passivity validity but only 11.1\% within 2~dB and .8074 reciprocity error, showing that inaccurate or collapsed predictions may still satisfy the passivity bound. After adaptation, passivity falls to 72.0--71.5\%, while response accuracy, reciprocity, and spectral consistency improve.

The task-specialized surrogate achieves the strongest pointwise accuracy, with .190 complex MAE and 84.4\% of predictions within 2~dB. However, only 28.3\% of its predicted frequencies satisfy passivity, and its reciprocity and spectral-derivative errors exceed the adapted models. The surrogate provides a strong numerical reference, whereas the adapted models offer a more balanced trade-off between response accuracy and physical consistency. Overall, T3 shows that pointwise accuracy, passivity, reciprocity, and spectral fidelity capture complementary aspects of EM response.

\paragraph{Response-to-design diagnosis and improvement (T5).}
T5 yields a different model ordering from the representation-focused tasks. GPT-5.4-mini achieves the strongest diagnosis performance, with 89.0\% accuracy and 88.4\% macro-F1, followed by AnalogSeeker at 85.0\% and 82.0\%, respectively. SFT improves the Qwen3-VL backbone from 51.0\% to 58.6\% macro-F1, whereas subsequent GRPO retains a smaller improvement at 55.5\%. These decision gains do not translate to physical redesign: end-to-end ADS improvement decreases from 48.4\% for the backbone to 45.3\% after SFT and 35.9\% after GRPO, while median ADS error reduction falls from 5.419~dB to 0.007 and 0.044~dB, respectively. This gap may reflect T5’s exclusion from SFT and GRPO’s surrogate-based rewards; it therefore indicates a gap between learning valid structured outputs and producing geometrically effective modifications under full-wave simulation. T5 exposes this diagnosis--redesign and reward--simulator mismatch, showing that strong performance on representation translation, structural understanding, and response prediction does not automatically yield reliable RF design improvement.

\paragraph{Structural and response understanding (T6 and T7).}
RFBench adaptation is highly effective for structural understanding. Qwen3 SFT achieves category macro-F1 scores of 99.4\% from primitives and 98.8\% from images, while GRPO reaches 99.3\% and 99.0\%. Their joint category-and-feature scores are similarly high. Thus, both models learn the procedural geometry families and associated structural properties.

Response classification in T7 improves substantially but remains less balanced. The strongest unadapted model, Claude~4.5, obtains 47.8\% accuracy and 36.4\% macro-F1. Qwen3 SFT and GRPO raise accuracy to 94.2\% and 94.3\%, respectively, but their macro-F1 scores remain 74.0\% and 74.2\%. The approximately 20-point accuracy--macro-F1 gap shows that aggregate accuracy overstates performance on the less frequent response categories and motivates reporting class-balanced metrics. Overall, SFT improves most tasks, while specification satisfaction and ADS-validated redesign remain challenging. GRPO yields modest, task-dependent changes; paired bootstrap intervals are reported in the supplement.

\paragraph{Limitations.}
RFBench currently focuses on simulated two-port layouts and does not model fabrication constraints or measured devices. Surrogate-based adaptation yields limited gains in ADS-validated T5 redesign, suggesting room for better alignment with full-wave simulation. Future work will broaden the RF structures and device settings and explore incorporating ADS feedback into the adaptation loop.

\section{Conclusion}

We introduce RFBench, a simulator-grounded dataset and seven-task benchmark that connects natural-language RF targets, editable layouts, images, occupancy grids, and electromagnetic responses. Experiments across five representative base models and two adapted Qwen3 variants demonstrate that RF-specific adaptation substantially improves layout validity, representation translation, EM-response prediction, structural understanding, and response classification. At the same time, closed-loop specification satisfaction and ADS-validated redesign remain challenging, revealing gaps that representation-level metrics alone cannot capture. By unifying multimodal representations with full-wave simulation, RFBench provides a rigorous foundation for evaluating and advancing models toward physically grounded RF design.

\bibliography{references}

\appendix
\setcounter{table}{0}
\setcounter{figure}{0}
\setcounter{equation}{0}
\renewcommand{\thetable}{S\arabic{table}}
\renewcommand{\thefigure}{S\arabic{figure}}
\renewcommand{\theequation}{S\arabic{equation}}
\setcounter{dbltopnumber}{4}
\renewcommand{\dbltopfraction}{0.95}
\renewcommand{\dblfloatpagefraction}{0.80}
\makeatletter
\setlength{\@dblfptop}{0pt}
\setlength{\@dblfpsep}{12pt}
\setlength{\@dblfpbot}{0pt plus 1fil}
\makeatother
\twocolumn[\begin{center}\Large\bf Supplementary Material\end{center}\vspace{1em}]
\raggedbottom
\section{Scope and Main-Paper Cross-Reference Map}

\begin{figure*}[t]
    \centering
    \includegraphics[width=0.92\textwidth]{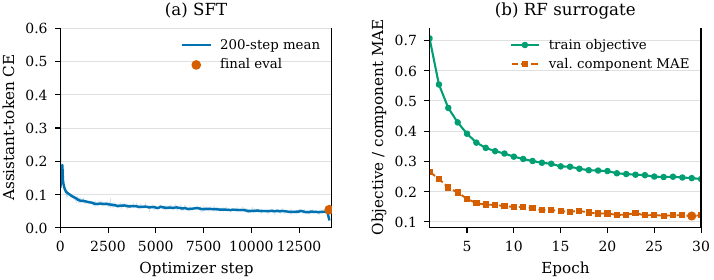}
    \caption{\textbf{Supervised optimization traces.} (a) Raw and 200-step-mean SFT cross entropy with final evaluation loss. (b) Surrogate training objective and held-out component MAE; the enlarged marker is the selected epoch.}
    \label{fig:training-diagnostics}
\end{figure*}

Task identifiers in this supplement follow the main paper exactly:
T1a, T1b, T2a, T2b, T3a, T3b, T4, T5, T6a, T6b, and T7.
The prefix ``S'' is reserved for supplementary figures and tables; it does not
denote a separate task family. The full-corpus distributions and task volumes
referenced in the main paper are Figure~\ref{fig:full-corpus} and
Table~\ref{tab:full-corpus}. The T3 component, magnitude, per-$S_{ij}$, and
threshold diagnostics are Tables~\ref{tab:t3-official-detail} and
\ref{tab:t3-official-perentry}. Adaptation settings and surrogate validation
are Table~\ref{tab:adaptation} and Figure~\ref{fig:training-diagnostics}.
The exploratory aggregate sensitivity and decomposition are
Tables~\ref{tab:rqs-sensitivity} and \ref{tab:rqs-decomposition}. Complete
task prompts and output schemas appear in the final section. The metric
definitions below cover every metric named in the main paper; all percentages
reported in either document are the corresponding rates multiplied by 100.

\section{Data Generation}

\subsection{EM template and binary representation}
In this work, we introduce five kinds of electromagnetic solved designs. Each design can be represented using a 2-D binary representation for its layout, assuming infinitely large, continuous ground plane underneath. Each layout can be represented by
\[
    \bm{X}\in\{0,1\}^{N\times N},\qquad N=16,
\]
where $X_{rc}=1$ denotes top-layer copper and $X_{rc}=0$ denotes an etched pixel.
The pixelized region was $20\,\mathrm{mm}\times20\,\mathrm{mm}$; hence
\[
    \Delta x=\Delta y=\frac{20}{16}=1.25\ \mathrm{mm}.
\]
It was embedded in a $30\,\mathrm{mm}\times24\,\mathrm{mm}$ two-port board with
$1.15\,\mathrm{mm}$-wide feeds.  The stack consisted of $35\,\mu\mathrm{m}$ copper on
Rogers RO4003C with
\[
    \varepsilon_r=3.55,\qquad \tan\delta=0.0027,\qquad h=0.508\ \mathrm{mm},
\]
and a continuous bottom ground plane.  The Momentum frequency plan was
\[
    f_m=\left(1+m\frac{9}{200}\right)\mathrm{GHz},
    \qquad m=0,\ldots,200.
\]

\subsection{Balanced mixture of geometry generators}

Let $K=5$ denote the number of geometry families,
\[
\mathcal{K}=\{\text{edge-connected},\ldots,\text{microstrip primitives}\}.
\]
The accepted-count target for family $k$ was
\[
    n_k=\left\lfloor\frac{10{,}000}{K}\right\rfloor
    +\1{k\leq 10{,}000\bmod K}=2000,
\]
so the final dataset contained exactly $2000$ accepted layouts from each family.
The pseudo-random generator was initialized with seed $7$.

For the edge-connected family, a left-to-right random-walk backbone
$\bm{B}$ was first constructed.  If $r_c$ is the backbone row at column $c$,
\begin{align*}
    r_0 &\sim \mathcal{U}_{\mathbb{Z}}\!\left(\left\lfloor N/3\right\rfloor,
    \left\lfloor 2N/3\right\rfloor-1\right),\\
    r_c &= \operatorname{clip}\!\left(r_{c-1}+\Delta r_c,1,N-2\right),\\
    \Pr(\Delta r_c=-1,0,+1)&=(0.25,0.50,0.25).
\end{align*}
The two vertical neighbors of each backbone pixel were independently activated with
probability $0.35$.  A sparse Bernoulli field was then added:
\begin{align*}
    p&\sim\mathcal{U}(0.08,0.25),&
    F_{rc}&\sim\operatorname{Bernoulli}(p),\\
    \bm{X}&=\bm{B}\lor\bm{F}.&&
\end{align*}
With probability $0.75$, only the largest four-neighbor connected component was retained.

The three symmetry-conditioned families were obtained from an edge-connected base
$\bm{X}_0$ using
\begin{align*}
    \bm{X}_{\mathrm{LR}} &= \mathcal{M}_{\mathrm{LR}}(\bm{X}_0),\\
    \bm{X}_{\mathrm{TB}} &= \mathcal{M}_{\mathrm{TB}}(\bm{X}_0),\\
    \bm{X}_{180} &= \bm{X}_0\lor\mathcal{R}_{180}(\bm{X}_0),
\end{align*}
where $\mathcal{M}_{\mathrm{LR}}$ and $\mathcal{M}_{\mathrm{TB}}$ mirror the retained
half-grid about the corresponding symmetry axis.

The microstrip-primitive family began with a centered trace of one or three pixel rows.
A random number of open-stub-like branches was then added:
\begin{align*}
    n_{\mathrm{stub}}\sim\mathcal{U}_{\mathbb{Z}}(2,6),\quad
    \ell_s\sim\mathcal{U}_{\mathbb{Z}}(2,7),\\
    w_s\sim\mathcal{U}_{\mathbb{Z}}(1,2),\quad
    d_s\in\{-1,+1\}.
\end{align*}

A local center-line gap was inserted with probability $0.55$, after which the largest
four-neighbor component was retained.

\subsection{Four-neighbor topology and launch constraints}

Define the copper fill ratio
\[
    \rho(\bm{X})=\frac{1}{N^2}\sum_{r=1}^{N}\sum_{c=1}^{N}X_{rc}.
\]
Connectivity was evaluated on the graph
\begin{align*}
    G(\bm{X})&=(V,E),\\
    V&=\{(r,c):X_{rc}=1\},\\
    E&=\{(u,v):\|u-v\|_1=1\}.
\end{align*}
Thus, diagonal contact alone did not establish electrical connectivity.  Let
$C_4(\bm{X})$ be the number of connected components of $G(\bm{X})$, and let
$I_{\mathrm{LR}}(\bm{X})$ indicate that one component intersects both feed-aligned
left and right boundaries.  For this grid, the feed overlaps rows
\[
    \mathcal{R}_{f}=\{7,8\}
\]
under zero-based indexing.  Before validation, launch pixels were forced to metal:
\[
    X_{r,0}=X_{r,N-1}=1,\qquad r\in\mathcal{R}_{f}.
\]

Let $D_{\diag}(\bm{X})$ count the $2\times2$ diagonal-only patterns
\[
    \begin{bmatrix}1&0\\0&1\end{bmatrix},\qquad
    \begin{bmatrix}0&1\\1&0\end{bmatrix}.
\]
A candidate was feasible only if its fill ratio lay in $[0.05,0.85]$, one four-neighbor
component joined the two feed boundaries, no diagonal-only block occurred, and the layout
was neither empty nor full. The configured minimum axis-aligned run was one pixel and was
therefore automatically satisfied by every nonempty layout.

\subsection{Penalty and rejection sampling}

The 10,000-sample generator used a \emph{hard feasibility
penalty}, rather than a continuously weighted score. Define
\begin{align*}
P_{\mathrm{geo}}
 &=P_{\rho}+P_{L}+P_{C}\notag\\
 &\quad+P_{D}+P_{F}+P_{T},
\end{align*}
where
\begin{align*}
P_{\rho}
 &=\1{\rho<0.05}
  +\1{\rho>0.85},\\
P_{L}
 &=\1{I_{\mathrm{LR}}=0},\\
P_{C}
 &=\1{C_4>1},\\
P_{D}
 &=\1{D_{\diag}>0},\\
P_{F}
 &=\sum_{r\in\mathcal{R}_f}
   \1{X_{r,0}X_{r,N-1}=0},\\
P_{T}
 &=\1{\lVert\bm{X}\rVert_1=0}
  +\1{\lVert\bm{X}\rVert_1=N^2}.
\end{align*}
The actual acceptance rule was
\[
    A(\bm{X})=\1{P_{\mathrm{geo}}(\bm{X})=0}
    =\mathcal{V}(\bm{X}).
\]
Equivalently, the barrier penalty was
\[
    \bar P_{\mathrm{geo}}(\bm{X})=
    \begin{cases}
      0, & P_{\mathrm{geo}}(\bm{X})=0,\\
      +\infty, & P_{\mathrm{geo}}(\bm{X})>0.
    \end{cases}
\]
Therefore, one violation was sufficient to reject the entire candidate; violations were
not averaged and could not compensate for one another.  If $q_k(\bm{X})$ is the proposal
distribution of family $k$, the accepted distribution was
\begin{align*}
    p_k(\bm{X}\mid\mathcal{V}=1)
    &=\frac{q_k(\bm{X})\mathcal{V}(\bm{X})}{Z_k},\\
    Z_k&=\sum_{\bm{X}}q_k(\bm{X})\mathcal{V}(\bm{X}).
\end{align*}
Candidates were repeatedly drawn until $n_k=2000$ feasible layouts were collected for
each family, with an upper bound of $1000n_k$ proposal attempts per family.  The five
accepted subsets were concatenated and randomly permuted.

The resulting index contained $10{,}000$ feasible layouts.  Its observed fill-ratio
statistics were
\begin{align*}
    \rho_{\min}&=0.085938,&\rho_{\max}&=0.414062,\\
    \bar\rho&=0.197465.&&
\end{align*}

\subsection{EM-ready conversion}

For each accepted layout, consecutive metal pixels in each row were merged into physical
rectangles.  The fixed feed rectangles were appended, and the sample was exported as
\[
    \left\{\bm{X}_i,\ \text{tokens}_i,\ \text{SVG}_i,
    \ \text{metadata}_i,\ \text{EM job}_i\right\}_{i=1}^{10{,}000}.
\]
The corresponding Momentum response is
\[
    \bm{S}_i(f_m)=\mathcal{M}_{\mathrm{EM}}\!\left(\bm{X}_i;
    \varepsilon_r,\tan\delta,h,t_{\mathrm{Cu}},\text{ports}\right),
\]
where $\mathcal{M}_{\mathrm{EM}}$ denotes the full-wave Momentum operator.

\section{Corpus Quality Control and Fixed Evaluation Panels}

\paragraph{Automated corpus checks.}
The retained corpus has 10,000 unique design identifiers and 10,000 distinct
primitive geometries; no exact geometry is duplicated. Every design uses the
same 201-point frequency grid from 1 to 10~GHz, and every stored real and
imaginary $S$-parameter value is finite. Recomputing
$20\log_{10}\max(|S_{ij}|,10^{-12})$ from the stored complex components differs
from the retained magnitude by at most $5.6\times10^{-6}$~dB. Rasterizing each
primitive geometry with the released converter reproduces its stored
$16\times16$ occupancy grid exactly. These checks are applied before task
construction, so all task views joined by design ID inherit the same geometry,
image, grid, response, structural category, and response label.

\paragraph{Response-label accounting.}
Each design has one fixed canonical response label stored before the
train--test split. Task builders for T1, T3, and T7 read that same
\texttt{response\_type} field and do not relabel a curve at evaluation time.
The full-corpus counts are 2,645 \texttt{broadband\_through}, 9
\texttt{broadband\_attenuating}, 2,697 \texttt{lowpass}, 372
\texttt{bandpass}, 1,499 \texttt{bandstop\_notch}, and 2,778
\texttt{resonant\_or\_shaped}, totaling 10,000. This fixed imbalance is why
T7 reports macro-F1 together with accuracy.

\paragraph{Split and panel integrity.}
The design split uses seed 42 and holds out exactly 200 designs from each of
the five structural families, for 1,000 official IDs; the other 9,000 designs
form the non-test pool. T1a, T1b, T2a, T2b, T3a, T3b, T6a, T6b, and T7 all
use these same IDs. T4 is a fixed 250-target subset with 50 targets from each
family. T5 uses a separate fixed list of 100 cases with 64 positive and 36
negative \emph{needs-improvement} labels. The seed-66 adaptation sampler and
the 8,100/900 surrogate train/validation partition are drawn only from
non-test IDs. The released panel verifier checks counts, joins, and exclusion
of evaluation IDs from every adaptation split.

\paragraph{Release scope.}
The dataset (\url{https://huggingface.co/datasets/Jerry1030/RFBench}) and code
(\url{https://github.com/jerry66s6/RFBench}) are released under CC BY~4.0 and
MIT, respectively. The release contains the full 10,000-design corpus with fixed
split identifiers, the held-out evaluation panels, task builders, exact prompt
templates, the evaluator, the predictions of every evaluated model, adaptation
configurations, and the frozen reward surrogate. ADS projects and proprietary
simulator software are not redistributed.

\section{Complete Main-Paper Metric Definitions}

Unless a metric is explicitly conditional, every official test example remains
in its denominator. Let $\1{\cdot}$ denote an indicator and let $N_t$ be the
official evaluation size for task $t$. Rates are first computed on $[0,1]$ and
then displayed as percentages. T3 numerical errors are the exception: as
specified below, they are evaluated over complete, evaluable complex sweeps.

\subsection{T1a/T1b: layout generation and category compliance}

For generated layout $n$, let $P_n$ indicate exactly two parsed port circles,
$M_n$ indicate that exactly two ports were parsed and both port-marker centers
lie on non-background metal, $C_n$ indicate that every parsed rectangle and
circle lies within the $30\,\mathrm{mm}\times24\,\mathrm{mm}$ canvas, and
$K_n$ indicate that one connected metal component touches both feed centers.
The main-paper structural-validity definition is
\begin{equation}
V_n=P_n C_n K_n
=\1{P_n=1\ \wedge\ C_n=1\ \wedge\ K_n=1}.
\label{eq:t1-validity}
\end{equation}
Thus ``On metal'' is a separately reported diagnostic and is not an additional
factor in $V_n$. For any binary layout diagnostic $Q_n\in\{P_n,M_n,C_n,K_n,V_n\}$,
\begin{equation}
\operatorname{Rate}(Q)=\frac{1}{N_1}\sum_{n=1}^{N_1}Q_n.
\label{eq:t1-rates}
\end{equation}

For T1b, let $\mathcal R_n$ be the nonempty set of requested category
constraints that the primitive evaluator can measure, and let $h_{n,r}=1$ when
constraint $r\in\mathcal R_n$ is satisfied. The per-layout mean measurable
compliance and full measurable compliance are
\begin{align}
c_n^{\rm meas}
&=\frac{1}{|\mathcal R_n|}\sum_{r\in\mathcal R_n}h_{n,r},\\
F_n^{\rm meas}
&=\prod_{r\in\mathcal R_n}h_{n,r}.
\label{eq:t1b-compliance}
\end{align}
Their reported panel values are $N_1^{-1}\sum_n c_n^{\rm meas}$ and
$N_1^{-1}\sum_n F_n^{\rm meas}$, respectively. ``Full spec.'' in the main
table is the latter rate. Generator texture or motif semantics that cannot be
executed from returned rectangles and circles are not silently treated as
failures; they are excluded from $\mathcal R_n$, as stated in
Table~\ref{tab:t1b-category}.

For completeness, Table~\ref{tab:t1-full10k} also reports two
supplementary-only generation diagnostics. If
$\mathcal B_n$ is the set of parsed non-background rectangles,
$A_n^{\rm sum}=\sum_{B\in\mathcal B_n}\operatorname{area}(B)$, and
$A_n^{\rm union}=\operatorname{area}(\bigcup_{B\in\mathcal B_n}B)$, then the
redundant-overlap diagnostic is
\begin{equation}
O_n=\begin{cases}
(A_n^{\rm sum}-A_n^{\rm union})/A_n^{\rm sum},&A_n^{\rm sum}>0,\\
0,&A_n^{\rm sum}=0.
\end{cases}
\label{eq:t1-supplemental-diagnostics}
\end{equation}
The table displays $N_1^{-1}\sum_n O_n$ as a percentage. Its ``Parse'' column
is $N_1^{-1}\sum_n q_n$, where $q_n$ is the number of primitive-parser errors
for output $n$; this is a mean count, not a percentage.

\subsection{T2a/T2b: binary-grid translation}

Let $L=256$, let $y_{n\ell}\in\{0,1\}$ be the target occupancy, and let
$u_n$ indicate that the prediction parses as exactly $L$ bits. Define
$\tilde y_{n\ell}$ as the parsed prediction when $u_n=1$ and as zero otherwise.
The foreground counts are pooled over the complete panel:
\begin{align}
\mathrm{TP}&=\sum_{n,\ell}\tilde y_{n\ell}y_{n\ell},&
\mathrm{FP}&=\sum_{n,\ell}\tilde y_{n\ell}(1-y_{n\ell}),\\
\mathrm{FN}&=\sum_{n,\ell}(1-\tilde y_{n\ell})y_{n\ell}.&
\label{eq:t2-counts}
\end{align}
The four foreground metrics are
\begin{align}
\mathrm{Precision}_{\rm fg}
&=\frac{\mathrm{TP}}{\mathrm{TP}+\mathrm{FP}},&
\mathrm{Recall}_{\rm fg}
&=\frac{\mathrm{TP}}{\mathrm{TP}+\mathrm{FN}},\\
F_{1,\rm fg}
&=\frac{2\mathrm{TP}}{2\mathrm{TP}+\mathrm{FP}+\mathrm{FN}},\\
\mathrm{IoU}_{\rm fg}
&=\frac{\mathrm{TP}}{\mathrm{TP}+\mathrm{FP}+\mathrm{FN}},
\label{eq:t2-overlap}
\end{align}
with a zero value when a displayed denominator is zero. The normalized Hamming
distance, exact-grid-match accuracy, and normalized metal-area difference are
\begin{align}
H&=\frac{1}{N_2}\sum_{n=1}^{N_2}
\left[(1-u_n)+\frac{u_n}{L}\sum_{\ell=1}^{L}
\1{\tilde y_{n\ell}\ne y_{n\ell}}\right],\\
A_{\rm exact}
&=\frac{1}{N_2}\sum_{n=1}^{N_2}
u_n\1{\sum_{\ell=1}^{L}\1{\tilde y_{n\ell}\ne y_{n\ell}}=0},\\
D_{\rm area}
&=\frac{1}{N_2L}\sum_{n=1}^{N_2}
\left|\sum_{\ell=1}^{L}\tilde y_{n\ell}
-\sum_{\ell=1}^{L}y_{n\ell}\right|.
\label{eq:t2-grid-diagnostics}
\end{align}
Consequently, a malformed grid receives Hamming distance one and exact-match
zero; its all-zero substitute is used for the pooled foreground counts and
area difference.

\subsection{T4: ADS execution and specification satisfaction}

Let $N_4=250$, $r_n=1$ when ADS returns a complete evaluable sweep, and
$K_0=3$ be the number of requested frequency landmarks. For a successful run,
let $a_{n,k}^{ij}$ be the ADS magnitude in dB and
$a_{n,k}^{ij,\star}$ the target. Define
\begin{equation}
e_{n,k}^{ij}=\left|a_{n,k}^{ij}-a_{n,k}^{ij,\star}\right|.
\label{eq:t4-landmark-error}
\end{equation}
The run rate, conditional per-entry landmark MAE, aggregate landmark MAE, and
combined reciprocal-entry diagnostic are
\begin{align}
R_{\rm run}^{(4)}
&=\frac{1}{N_4}\sum_{n=1}^{N_4}r_n,\\
\operatorname{MAE}_{ij}^{(4)}
&=\frac{\sum_{n=1}^{N_4}r_n\sum_{k=1}^{K_0}e_{n,k}^{ij}}
{K_0\sum_{n=1}^{N_4}r_n},\\
\operatorname{MAE}_{\rm all}^{(4)}
&=\frac{1}{4}\sum_{i,j=1}^{2}\operatorname{MAE}_{ij}^{(4)},\\
\operatorname{MAE}_{12=21}^{(4)}
&=\frac{\operatorname{MAE}_{12}^{(4)}
+\operatorname{MAE}_{21}^{(4)}}{2}.
\label{eq:t4-run-mae}
\end{align}
The strict main-paper specification-success rate retains every target,
including failed simulations, in its denominator:
\begin{equation}
\begin{aligned}
R_{\rm strict,3dB}^{(4)}
&=\frac{1}{N_4}\sum_{n=1}^{N_4}r_n I_n^{(3)},\\
I_n^{(3)}
&=\prod_{i,j=1}^{2}\prod_{k=1}^{K_0}
\1{e_{n,k}^{ij}\leq3\ {\rm dB}}.
\end{aligned}
\label{eq:t4-strict}
\end{equation}

\subsection{Classification metrics shared by T6 and T7}

For a $C$-class problem with ground truth $y_n$ and prediction $\hat y_n$,
define
\begin{align}
\mathrm{TP}_c&=\sum_n\1{y_n=c,\hat y_n=c},\\
\mathrm{FP}_c&=\sum_n\1{y_n\ne c,\hat y_n=c},\\
\mathrm{FN}_c&=\sum_n\1{y_n=c,\hat y_n\ne c}.
\end{align}
Then
\begin{align}
\mathrm{Accuracy}
&=\frac{1}{N}\sum_{n=1}^{N}\1{\hat y_n=y_n},\\
\mathrm{Precision}_c
&=\frac{\mathrm{TP}_c}{\mathrm{TP}_c+\mathrm{FP}_c},\\
\mathrm{Recall}_c
&=\frac{\mathrm{TP}_c}{\mathrm{TP}_c+\mathrm{FN}_c},\\
F_{1,c}
&=\frac{2\mathrm{TP}_c}
{2\mathrm{TP}_c+\mathrm{FP}_c+\mathrm{FN}_c},\\
\mathrm{MacroF1}
&=\frac{1}{C}\sum_{c=1}^{C}F_{1,c},
\label{eq:classification-metrics}
\end{align}
with zero assigned when a class-specific denominator is zero. Invalid labels
are incorrect predictions and remain in $N$. For T6, $C=5$ structural
categories; for T7, $C=6$ response types.

\subsection{T6a/T6b: category and feature understanding}

Let $\mathcal J$ contain the nine primary non-diagonal feature flags, so
$J=|\mathcal J|=9$, and let
$z_{n,j}=\1{\hat f_{n,j}=f_{n,j}}$. Category accuracy and category macro-F1
use Equation~\ref{eq:classification-metrics}. The remaining main-paper
metrics are
\begin{align}
A_j&=\frac{1}{N_6}\sum_{n=1}^{N_6}z_{n,j},\\
\bar A_{\rm feat}
&=\frac{1}{N_6J}\sum_{n=1}^{N_6}\sum_{j\in\mathcal J}z_{n,j},\\
A_{\rm exact\text{-}feat}
&=\frac{1}{N_6}\sum_{n=1}^{N_6}\prod_{j\in\mathcal J}z_{n,j},\\
A_{\rm joint}
&=\frac{1}{N_6}\sum_{n=1}^{N_6}
\1{\hat y_n=y_n}\prod_{j\in\mathcal J}z_{n,j}.
\label{eq:t6-feature-metrics}
\end{align}
These four lines are, respectively, per-feature accuracy, mean
primary-feature accuracy, exact feature-vector match, and joint
category-plus-feature correctness.
The tenth flag, \texttt{diagonal\_connected\_blocks}, is retained as a separate
diagnostic with accuracy given by the first line, but it is excluded from
$\bar A_{\rm feat}$, exact feature match, and joint correctness.

\subsection{T7: six-way RF-response classification}

Six-way accuracy and macro-F1 are Equation~\ref{eq:classification-metrics}
with the six response labels listed in the main paper and in the prompt schema
below. Because all six class-specific $F_1$ values receive equal weight,
macro-F1 does not inherit the corpus class frequencies.

\section{Implemented GRPO Reward Function}

The reported run calls the released \texttt{compute\_score} router once for each sampled
completion. It does not use a mask-hinge objective or invoke ADS during rollout. Every task
branch returns a bounded score in $[0,1]$, so the four completions for one prompt can be
ranked on a common scale while retaining task-specific semantics.

\paragraph{Geometry and specification terms.}
T1a and positive T5 redesigns are parsed into rectangles and port circles, then rasterized on
the official $16\!\times\!16$ grid. Let $n$ be occupied cells, $\gamma$ the largest
four-neighbor component fraction, $e$ the number of occupied feed boundaries, and $p=2$
when both required port markers are present ($p=0$ when circles are omitted). The exact
graded geometry term is
\begin{align*}
v={}&0.25+0.20\1{4\le n\le180}\\
&+0.20\1{\gamma\ge.65}+0.10\1{.40\le\gamma<.65}\\
&+0.20\1{e=2}+0.08\1{e=1}\\
&+0.15\1{p=2}+0.05\1{p=0},
\end{align*}
clipped to $[0,1]$. Empty, almost-full ($n>230$), or unparsable geometry has $v=0$;
geometry is valid only when $v\ge0.55$ and metal reaches both feed boundaries.

For valid candidates, the frozen CNN predicts real and imaginary values for all four
$S_{ij}$ entries at 201 frequencies. With $a_{ij}(f)=20\log_{10}|S_{ij}(f)|$, its requested
landmark score is
\[
p_S=\operatorname{mean}_{k}\operatorname{clip}\!\left(
1-\frac{|\hat a_{i_kj_k}(f_k)-a_k^\star|}{20\ \mathrm{dB}},0,1\right).
\]
The term $p_H$ averages clipped errors for requested minimum, peak, and median $S_{21}$
magnitude (15-dB scale) and requested characteristic frequencies or widths (2-GHz scale).
If a response class is supplied, $p_C$ is exact class correctness. The surrogate
specification term is
\[
z=\begin{cases}
0.65p_S+0.25p_H+0.10p_C,&\text{class supplied},\\
0.70p_S+0.30p_H,&\text{otherwise}.
\end{cases}
\]
The selected surrogate uses 8,100 non-test training designs and 900 disjoint non-test
validation designs; neither split overlaps official evaluation IDs.

\paragraph{Single routed reward.}
Let $b$ be the T5 ground-truth need for improvement, $\hat b$ the parsed yes/no decision,
$V$ redesign validity, $z_0$ the supplied current response scored against the same target,
and $d=\operatorname{clip}(0.5+z-z_0,0,1)$. The complete T5 branch is
\[
\rho_5=\begin{cases}
0,&\text{unparsable output},\\
1.00,&(b,\hat b)=(0,0),\\
0.05,&(b,\hat b)=(1,0),\\
0.10,&(b,\hat b)=(0,1),\\
0.10+0.15v,&(b,\hat b,V)=(1,1,0),\\
0.20v+0.55z+0.25d,&(b,\hat b,V)=(1,1,1).
\end{cases}
\]
Writing $F_{\rm fg}$ for foreground F1, $E_g$ for exact-grid correctness, $m$ for
real/imaginary component MAE, $c$ for category correctness, $\bar f$ for mean feature
correctness, and $q$ for class correctness, the released router is the single function
\begin{equation}
R(\tau,y)=\begin{cases}
0.15v,&\tau=\mathrm{T1a},\ \text{invalid},\\
0.25v+0.75z,&\tau=\mathrm{T1a},\ \text{valid},\\
0.75F_{\rm fg}+0.25E_g,&\tau=\mathrm{T2a/b},\\
\operatorname{clip}(1-m/0.25,0,1),&\tau=\mathrm{T3}_{\mathrm{comp}},\\
\rho_5,&\tau=\mathrm{T5},\\
0.70c+0.30\bar f,&\tau=\mathrm{T6a/b},\\
q,&\tau\in\{\mathrm{T3}_{\mathrm{type}},\mathrm{T7}\},\\
0,&\text{otherwise}.
\end{cases}
\label{eq:reward-router}
\end{equation}
T1b and T4 are absent from the GRPO pool. This staging first tests usable syntax and
geometry, then supplies dense RF ranking for valid designs. The surrogate is an optimization
signal: official MLLM metrics never use its
predictions as targets. T4 evaluates the same
predeclared class of all-$S_{ij}$ landmarks using ADS, making its SFT--GRPO comparison an
out-of-reward-model check of the intended shaping signal.

\section{T5 Paired ADS Improvement Metrics}

T5 is evaluated in two stages on the same fixed $N_5=100$ case panel. Stage A
scores the binary decision on every case. Let $b_n\in\{0,1\}$ be the
ground-truth need for improvement and let $\hat b_n$ be the parsed model
decision. For class $c\in\{0,1\}$, define
\begin{align}
\mathrm{TP}^{(5)}_c
&=\sum_{n=1}^{N_5}\1{b_n=c,\hat b_n=c},\\
\mathrm{FP}^{(5)}_c
&=\sum_{n=1}^{N_5}\1{b_n\ne c,\hat b_n=c},\\
\mathrm{FN}^{(5)}_c
&=\sum_{n=1}^{N_5}\1{b_n=c,\hat b_n\ne c}.
\end{align}
The Stage-A metrics are
\begin{align}
\mathrm{Accuracy}^{(5)}
&=\frac{1}{N_5}\sum_{n=1}^{N_5}\1{\hat b_n=b_n},\\
\mathrm{Precision}^{(5)}_{+}
&=\frac{\mathrm{TP}^{(5)}_1}
{\mathrm{TP}^{(5)}_1+\mathrm{FP}^{(5)}_1},\\
\mathrm{Recall}^{(5)}_{+}
&=\frac{\mathrm{TP}^{(5)}_1}
{\mathrm{TP}^{(5)}_1+\mathrm{FN}^{(5)}_1},\\
F^{(5)}_{1,c}
&=\frac{2\mathrm{TP}^{(5)}_c}
{2\mathrm{TP}^{(5)}_c+\mathrm{FP}^{(5)}_c+\mathrm{FN}^{(5)}_c},\\
\mathrm{MacroF1}^{(5)}
&=\frac{1}{2}\sum_{c\in\{0,1\}}F^{(5)}_{1,c}.
\label{eq:t5-stage-a}
\end{align}
The positive class is \emph{needs improvement}. Invalid decisions are
incorrect and remain in the 100-case denominator. For a Stage-A true positive,
the proposed redesign is independently solved with ADS Momentum and paired
with the supplied original design.
Let $K_n$ be the number of requested frequency landmarks for case $n$, let
$a^{ij,\star}_{n,k}$ be the target magnitude in dB for entry $S_{ij}$ at landmark
frequency $f_{n,k}$, and let $a^{ij}_{n,t,k}$ be the corresponding ADS-simulated
magnitude for the original ($t=0$) or redesigned ($t=1$) geometry. The all-$S$
specification error is
\begin{equation}
E_{n,t}=\frac{1}{4K_n}
\sum_{k=1}^{K_n}\sum_{i,j=1}^{2}
\left|a^{ij}_{n,t,k}-a^{ij,\star}_{n,k}\right|,
\qquad t\in\{0,1\}.
\label{eq:t5-spec-error}
\end{equation}
Thus, the paired error change is defined as
\begin{equation}
\Delta E_n=E_{n,0}-E_{n,1},
\label{eq:t5-delta-error}
\end{equation}
so that $\Delta E_n>0$, equivalently $E_{n,1}<E_{n,0}$, denotes an improved
redesign.

Using the same $b_n$ and $\hat b_n$, let $V_n$ indicate redesign validity and
let $A_n$ indicate that the ADS solve completed and produced an evaluable
response. Define the true-positive candidate, successfully solved, and
improved sets as
\begin{align}
\mathcal P&=\{n:b_n=1\},\\
\mathcal C&=\{n:b_n=1,\ \hat b_n=1\},\\
\mathcal S&=\{n\in\mathcal C:V_n=1,\ A_n=1\},\\
\mathcal I&=\{n\in\mathcal S:E_{n,1}<E_{n,0}\}.
\label{eq:t5-eval-sets}
\end{align}
The reported ADS run rate, conditional improvement rate, and median paired
target-error reduction are
\begin{align}
R_{\rm run}^{(5)}
&=\frac{|\mathcal S|}{|\mathcal C|}
&&\text{(solved/candidates)},\\
R_{\rm improve\mid solved}^{(5)}
&=\frac{|\mathcal I|}{|\mathcal S|}
&&\text{(improved/solved)},\\
\widetilde{\Delta E}
&=\operatorname*{median}_{n\in\mathcal S}\Delta E_n.
\label{eq:t5-conditional-metrics}
\end{align}
The end-to-end ADS improvement rate instead retains every
ground-truth-positive case in its denominator:
\begin{equation}
R_{\mathrm{E2E}}
=\frac{|\mathcal I|}{|\mathcal P|}
=\frac{|\mathcal I|}{\sum_n\1{b_n=1}}.
\label{eq:t5-e2e-improvement}
\end{equation}
The denominator is 64 on the official panel. Consequently, false-negative
decisions, invalid redesigns, ADS failures, and solved redesigns with
$E_{n,1}\ge E_{n,0}$ all contribute zero to this end-to-end rate. The reported
median paired change therefore contains no failed solve.

\section{T3 Metric Definitions and Official-Panel Diagnostics}

Let $\hat{\bm S}_{n,k}$ and $\bm S_{n,k}$ denote the predicted and ADS-reference
$2\times2$ complex scattering matrices for complete, evaluable design
$n\in\{1,\ldots,N_3\}$ at the $k$th of $F=201$ frequencies. The component
errors and main complex MAE are
\begin{align}
\operatorname{MAE}_{\rm Re}
&=\frac{1}{4N_3F}\sum_{n=1}^{N_3}\sum_{i,j=1}^{2}\sum_{k=1}^{F}
\left|\operatorname{Re}\hat S_{n,k}^{ij}
-\operatorname{Re}S_{n,k}^{ij}\right|,\\
\operatorname{MAE}_{\rm Im}
&=\frac{1}{4N_3F}\sum_{n=1}^{N_3}\sum_{i,j=1}^{2}\sum_{k=1}^{F}
\left|\operatorname{Im}\hat S_{n,k}^{ij}
-\operatorname{Im}S_{n,k}^{ij}\right|,
\label{eq:t3-components}
\end{align}
and
\begin{equation}
E_{\mathrm{complex}}=
\frac{1}{4N_3F}\sum_{n=1}^{N_3}\sum_{i,j=1}^{2}\sum_{k=1}^{F}
\left|\hat S_{n,k}^{ij}-S_{n,k}^{ij}\right|,
\label{eq:t3-complex}
\end{equation}
respectively. Define
$a^{ij}_{n,k}=20\log_{10}\max(|S^{ij}_{n,k}|,10^{-12})$ and its predicted
counterpart $\hat a^{ij}_{n,k}$, with
$d^{ij}_{n,k}=|\hat a^{ij}_{n,k}-a^{ij}_{n,k}|$. The overall and per-entry
magnitude metrics are
\begin{align}
\operatorname{MAE}_{\rm dB}
&=\frac{1}{4N_3F}\sum_{n,i,j,k}d^{ij}_{n,k},\\
\operatorname{RMSE}_{\rm dB}
&=\sqrt{\frac{1}{4N_3F}\sum_{n,i,j,k}(d^{ij}_{n,k})^2},\\
\operatorname{MAE}_{\rm dB}^{ij}
&=\frac{1}{N_3F}\sum_{n=1}^{N_3}\sum_{k=1}^{F}d^{ij}_{n,k}.
\label{eq:t3-magnitude-errors}
\end{align}
Here and in this paragraph, $\sum_{n,i,j,k}$ spans
$n=1,\ldots,N_3$, $i,j\in\{1,2\}$, and $k=1,\ldots,F$. The within-$\delta$-dB
point rate is
\begin{equation}
R_{\leq\delta{\rm dB}}
=\frac{1}{4N_3F}\sum_{n=1}^{N_3}\sum_{i,j=1}^{2}\sum_{k=1}^{F}
\1{d^{ij}_{n,k}\leq\delta}.
\label{eq:t3-within-threshold}
\end{equation}
We retain $\delta\in\{0.5,1,2\}$ in the released evaluator and use $2$~dB in
the main comparison.

For every T3 row in the main paper, all 1,000 predictions pass the complete
sweep check, so $N_3=1000$ and the component, complex, magnitude, and
within-threshold metrics each use exactly
$4N_3F=4\times1000\times201=804{,}000$ matrix--frequency values. The
complete-sweep filter is retained as an evaluator safeguard.

\paragraph{Passivity.}
For each predicted matrix, the evaluator computes its largest singular value. Under the
common power-wave normalization used here, passivity requires
$\sigma_{\max}(\hat{\bm S}_{n,k})\leq1+\epsilon_{\rm ADS}$. The tolerance is not
chosen manually; it is the maximum positive residual on the official ADS ground truth:
\begin{align}
\epsilon_{\rm ADS}
&=\max_{n,k}\max\!\left(0,\sigma_{\max}(\bm S_{n,k})-1\right)=0,\\
\max_{n,k}\sigma_{\max}(\bm S_{n,k})&=0.997162.
\end{align}
The reported rate and optional mean violation are
\begin{align}
\mathrm{PassivityRate}
&=\frac{1}{N_3F}\sum_{n=1}^{N_3}\sum_{k=1}^{F}
\1{\sigma_{\max}(\hat{\bm S}_{n,k})\leq1+\epsilon_{\rm ADS}},\\
E_{\rm pass}
&=\frac{1}{N_3F}\sum_{n=1}^{N_3}\sum_{k=1}^{F}
\max\!\left(0,\sigma_{\max}^{2}(\hat{\bm S}_{n,k})-1\right).
\label{eq:t3-passivity}
\end{align}

\paragraph{Reciprocity and frequency evolution.}
The ADS export provenance records the same real $50\,\Omega$ reference impedance for both
ports, and the official reference curves have maximum complex
$|S_{12}-S_{21}|$ residual exactly zero. No port renormalization is therefore applied, and
complex reciprocity is measured as
\begin{equation}
E_{\rm rec}=\frac{1}{N_3F}\sum_{n=1}^{N_3}\sum_{k=1}^{F}
\left|\hat S^{12}_{n,k}-\hat S^{21}_{n,k}\right|.
\label{eq:t3-reciprocity}
\end{equation}
This simple equality would require modification for unequal or complex reference
impedances. To avoid rewarding a constant prediction merely for being smooth, spectral
continuity is measured relative to the reference changes. With
$\Delta S^{ij}_{n,k}=S^{ij}_{n,k+1}-S^{ij}_{n,k}$ (and analogously for
$\Delta\hat S$),
\begin{equation}
E_{\Delta}=\frac{1}{4N_3(F-1)}
\sum_{n=1}^{N_3}\sum_{i,j=1}^{2}\sum_{k=1}^{F-1}
\left|\Delta\hat S^{ij}_{n,k}-\Delta S^{ij}_{n,k}\right|.
\label{eq:t3-derivative}
\end{equation}
We make no causality, lossless energy-conservation, or magnitude--phase-consistency claim.

\paragraph{Task-specialized grid reference.}
The frozen CNN takes one geometry layout and emits tensors of shape
$2\times201\times2\times2$, with separate real and imaginary channels. It was trained on
8,100 non-test designs, selected by real/imaginary component MAE on 900 disjoint non-test
designs, and evaluated once on the unchanged official 1,000-design list. Its training loss is
\begin{equation}
\begin{aligned}
\mathcal L_{\rm sur}
={}&\operatorname{SmoothL1}_{\beta=.05}(\hat{\bm S},\bm S)\\
&+0.1\,\operatorname{SmoothL1}_{\beta=.05}(\hat{\bm A},\bm A),
\end{aligned}
\label{eq:surrogate-loss}
\end{equation}
where $\bm A=20\log_{10}\max(|\bm S|,10^{-12})$. The checkpoint and split are unchanged
from GRPO reward shaping. Its official predictions are evaluated by the same T3 evaluator
as all other rows, but the canonical grid input and single-task objective make it a
task-specialized reference rather than a directly comparable MLLM.

\begin{table*}[t]
\centering
\small
\setlength{\tabcolsep}{3.0pt}
\begin{tabular}{@{}llrrrrrrr@{}}
\toprule
Model & Input & Real MAE $\downarrow$ & Imag. MAE $\downarrow$ & dB MAE $\downarrow$ & dB RMSE $\downarrow$ & $\leq0.5$ dB $\uparrow$ & $\leq1$ dB $\uparrow$ & $E_{\rm pass}$ $\downarrow$ \\
\midrule
AnalogSeeker & primitive & .720 & .402 & 111.37 & 159.14 & 5.3 & 7.9 & .00213 \\
Claude 4.5 & primitive & .618 & .499 & 6.54 & 9.13 & 8.9 & 16.1 & .15933 \\
 & image & .622 & .484 & 6.76 & 9.64 & 8.5 & 15.6 & .18334 \\
GPT-5.4-mini & primitive & .594 & .417 & 9.82 & 14.10 & 8.3 & 16.6 & .00946 \\
 & image & .476 & .432 & 13.35 & 17.34 & 3.0 & 5.9 & .05732 \\
Llama 3.2-11B & primitive & .406 & .404 & 101.63 & 140.26 & 0.1 & 0.1 & .00000 \\
 & image & .440 & .403 & 80.06 & 126.32 & 0.8 & 1.6 & 1.11839 \\
Qwen3-VL-8B & primitive & .639 & .402 & 21.35 & 26.92 & 6.8 & 11.4 & .31399 \\
 & image & .523 & .403 & 32.86 & 53.17 & 3.8 & 7.0 & .00400 \\
Qwen3 SFT & primitive & .196 & .191 & 3.79 & 9.22 & 33.3 & 48.0 & .00995 \\
 & image & .187 & .184 & 3.54 & 7.93 & 35.0 & 49.5 & .01046 \\
Qwen3 GRPO & primitive & .196 & .190 & 3.72 & 8.58 & 33.3 & 48.0 & .01044 \\
 & image & .186 & .184 & 3.52 & 7.83 & 34.9 & 49.5 & .01025 \\
\midrule
Grid surrogate & grid & .122 & .110 & 1.36 & 3.23 & 54.5 & 71.9 & .13029 \\
\bottomrule
\end{tabular}
\caption{\textbf{Detailed T3 diagnostics on the official 1,000-design panel.} Threshold entries are percentages over all four matrix entries and 201 frequencies. Main-paper complex MAE, within-2-dB, passivity-rate, reciprocity, and spectral-derivative values come from the same evaluator output.}
\label{tab:t3-official-detail}
\end{table*}

\begin{table*}[t]
\centering
\small
\setlength{\tabcolsep}{4.2pt}
\begin{tabular}{@{}llrrrr@{}}
\toprule
Model & Input & $S_{11}$ dB MAE $\downarrow$ & $S_{12}$ dB MAE $\downarrow$ & $S_{21}$ dB MAE $\downarrow$ & $S_{22}$ dB MAE $\downarrow$ \\
\midrule
AnalogSeeker & primitive & 6.52 & 219.94 & 212.51 & 6.51 \\
Claude 4.5 & primitive & 8.11 & 4.97 & 4.97 & 8.11 \\
 & image & 7.99 & 6.11 & 4.97 & 7.99 \\
GPT-5.4-mini & primitive & 10.93 & 13.52 & 3.80 & 11.03 \\
 & image & 8.03 & 21.96 & 15.11 & 8.31 \\
Llama 3.2-11B & primitive & 66.07 & 134.15 & 143.89 & 62.41 \\
 & image & 77.96 & 83.64 & 80.34 & 78.32 \\
Qwen3-VL-8B & primitive & 22.81 & 36.13 & 4.00 & 22.46 \\
 & image & 29.55 & 50.00 & 22.22 & 29.65 \\
Qwen3 SFT & primitive & 4.01 & 3.58 & 3.58 & 3.97 \\
 & image & 3.84 & 3.25 & 3.25 & 3.81 \\
Qwen3 GRPO & primitive & 4.01 & 3.45 & 3.45 & 3.99 \\
 & image & 3.86 & 3.20 & 3.20 & 3.82 \\
\midrule
Grid surrogate & grid & 1.41 & 1.29 & 1.31 & 1.45 \\
\bottomrule
\end{tabular}
\caption{\textbf{Per-entry T3 magnitude errors} on the official 1,000-design panel. These detailed diagnostics are supplementary; the main comparison uses the complex and physics-aware metrics defined above.}
\label{tab:t3-official-perentry}
\end{table*}

\section{Full-Corpus Statistics and Task Availability}

\begin{figure*}[t]
    \centering
    \includegraphics[width=0.92\textwidth]{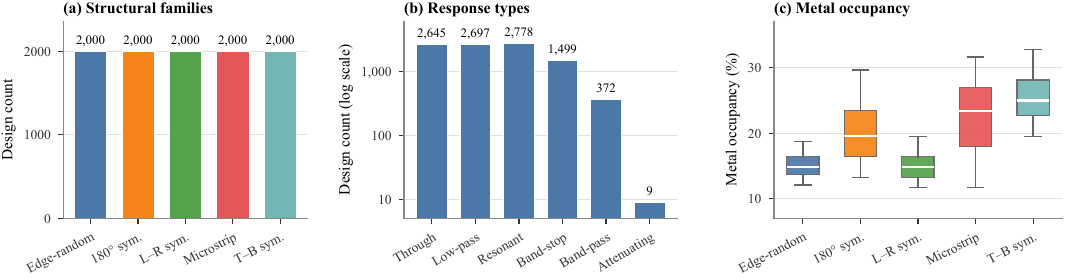}
    \caption{\textbf{Statistics of the 10K corpus.} (a) Each of the five structural families contains 2,000 EM Momentum designs. (b) Response labels derived from $S_{21}$ are naturally long-tailed. (c) The five generators induce distinct metal-occupancy distributions (boxes show the interquartile range; whiskers show the 5th--95th percentiles).}
    \label{fig:full-corpus}
\end{figure*}

\begin{table*}[t]
\centering
\small
\setlength{\tabcolsep}{4.2pt}
\begin{tabular}{@{}llllrr@{}}
\toprule
ID & Ability & Input & Output & Full corpus & Official test \\
\midrule
T1a & Design & RF response specification & Primitive layout & 10,000 & 1,000 \\
T1b & Constrained design & RF specification + category & Primitive layout & 10,000 & 1,000 \\
T2a & Representation translation & Primitive layout & $16\!\times\!16$ binary token & 10,000 & 1,000 \\
T2b & Representation translation & Layout image & $16\!\times\!16$ binary token & 10,000 & 1,000 \\
T3a & Response prediction & Primitive layout & Complex $2\!\times\!2$ $S$-parameter sweep & 10,000 & 1,000 \\
T3b & Response prediction & Layout image & Complex $2\!\times\!2$ $S$-parameter sweep & 10,000 & 1,000 \\
T4 & Simulator execution & T1a-generated layout & Run/fail + simulated response & -- & 250/model \\
T5 & Improvement diagnosis & Layout + response + target & Decision + proposed redesign & 500 & 100 \\
T6a & Structural understanding & Primitive layout & Category + 10 feature flags & 10,000 & 1,000 \\
T6b & Structural understanding & Layout image & Category + 10 feature flags & 10,000 & 1,000 \\
T7 & Response classification & Complex $S$-parameter sweep & One of six response types & 10,000 & 1,000 \\
\midrule
\multicolumn{4}{r}{Common-set task instances (T4 reported separately)} & 90,500 & 9,100 \\
\bottomrule
\end{tabular}
\caption{Full-corpus task availability versus the fixed official comparison set. Multiple task views may share one source design and remain aligned by design ID.}
\label{tab:full-corpus}
\end{table*}

The complete corpus contains 10,000 EM Momentum designs, exactly 2,000 in each structural family. Figure~\ref{fig:full-corpus} reports full-corpus response-label and metal-occupancy distributions. The response labels are intentionally imbalanced: only 9 designs are broadband attenuating and 372 are band-pass, whereas the other four classes contain 1,499--2,778 examples. The main paper therefore reports macro-F1 alongside accuracy.

The main paper reports model comparisons on a fixed official evaluation set: 1,000 shared designs for T1a/T1b/T2/T3/T6/T7, a category-balanced 250-design subset for closed-loop T4, and 100 shared diagnosis cases for T5. Table~\ref{tab:full-corpus} records the larger task views that can be constructed from the complete corpus. These counts describe data availability, not the number of examples used to compute the reported model scores.

\paragraph{Structural-family semantics.}
The five generator labels denote visibly and algorithmically distinct construction rules. \texttt{edge\_connected\_random} contains feed-to-feed connected, random-walk-like metal with stochastic fill. \texttt{microstrip\_primitives} uses an RF-style center trace with optional stubs or resonator-like branches. \texttt{rot180} enforces 180-degree rotational symmetry about the layout center. \texttt{symmetric\_lr} and \texttt{symmetric\_tb} enforce mirror symmetry about the vertical and horizontal center lines, respectively. These labels are generator metadata. The $16\!\times\!16$ binary token marks occupied metal cells; the occupancy in Figure~\ref{fig:full-corpus}(c) is therefore the percentage of its 256 cells equal to one.

\paragraph{Response-label semantics.}
The six T7 labels summarize the full $S_{21}$ sweep. \texttt{broadband\_through} denotes comparatively low-loss, smooth transmission across the band, while \texttt{broadband\_attenuating} denotes weak transmission across most of it. \texttt{lowpass} requires stronger low-frequency transmission followed by high-frequency roll-off. \texttt{bandpass} contains a localized transmission peak with weaker transmission on both sides, and \texttt{bandstop\_notch} contains localized rejection with better-transmitting neighbors. \texttt{resonant\_or\_shaped} is the residual non-flat or resonant class when the response does not cleanly match the other five. Each label is stored with the complete simulated sweep before benchmark splitting and is shared by T1a, T3, and T7 through the design ID; evaluation predictions are joined only after the fixed ground-truth files are built.

\section{Category-Level and Full-Corpus Model Results}

Table~\ref{tab:t1b-category} decomposes T1b by the five fixed generator categories.
Each cell reports mean measurable-constraint compliance followed by the stricter rate at
which every measurable constraint is satisfied. The table is computed on the official
balanced panel, with 200 targets per category.

\begin{table*}[t]
\centering
\small
\setlength{\tabcolsep}{4.2pt}
\begin{tabular}{@{}lrrrrr@{}}
\toprule
Model & Edge-connected & Microstrip & Rot180 & Symmetric L--R & Symmetric T--B \\
\midrule
AnalogSeeker & 0.0/0.0 & 0.0/0.0 & 50.0/2.0 & 80.0/73.5 & 0.0/0.0 \\
Claude 4.5 & 0.0/0.0 & 0.0/0.0 & 98.8/97.5 & 98.5/97.0 & 44.3/0.0 \\
GPT-5.4-mini & 0.0/0.0 & 0.0/0.0 & 0.0/0.0 & 0.0/0.0 & 0.0/0.0 \\
Llama 3.2-11B & 0.0/0.0 & 0.0/0.0 & 1.0/0.0 & 3.5/0.5 & 6.0/0.0 \\
Qwen3-VL-8B & 0.0/0.0 & 0.0/0.0 & 19.3/0.0 & 18.0/4.0 & 0.3/0.0 \\
Qwen3 SFT & 58.5/58.5 & \textbf{100.0/100.0} & 85.5/80.0 & 97.5/95.0 & 97.5/95.0 \\
Qwen3 GRPO & 55.0/55.0 & \textbf{100.0/100.0} & \textbf{86.8}/79.5 & \textbf{97.8/95.5} & \textbf{98.0/96.0} \\
\bottomrule
\end{tabular}
\caption{\textbf{T1b per-category compliance} on the official 1,000-design panel (percent). Each entry is mean/full measurable compliance. Texture semantics that are not executable from generated primitives are excluded from both denominators, as in the aggregate T1b metric.}
\label{tab:t1b-category}
\end{table*}

The following tables mirror the native metric schemas in the main paper while replacing the
official panel by the largest archived evaluation panel for each task. T1a/T1b, T2, T6, and
T7 have complete 10,000-design outputs for the five unadapted models. T3 has complete 10K
outputs only for AnalogSeeker-a, Llama 3.2-11B-a/b, and Qwen3-VL-8B-a/b; other T3 outputs contain
only the official 1,000 designs. SFT and GRPO likewise have only the official 1,000-design
outputs and are therefore not repeated in the 10K tables. T4 is a closed-loop experiment
defined on 250 targets, and T5 contains 500 full-pool cases. We report
their complete available panels separately instead of extrapolating or relabeling them as
10K. The companion \texttt{full\_10000\_all\_metrics.csv} preserves all 38 complete-run
summary rows and all 73 evaluator columns.

\begin{table*}[t]
\centering
\small
\setlength{\tabcolsep}{3.4pt}
\begin{tabular}{@{}llrrrrrrrrr@{}}
\toprule
Model & ID & Ports $\uparrow$ & On metal $\uparrow$ & Canvas $\uparrow$ & Connect $\uparrow$ & Valid $\uparrow$ & Redund. overlap $\downarrow$ & Parse $\downarrow$ & Meas. mean $\uparrow$ & Meas. full $\uparrow$ \\
\midrule
AnalogSeeker & T1a & 98.1 & 77.7 & 0.0 & 0.0 & 0.0 & 10.9 & .012 & -- & -- \\
Claude 4.5 & T1a & 100.0 & 88.7 & 0.0 & 0.0 & 0.0 & 2.5 & .000 & -- & -- \\
GPT-5.4-mini & T1a & 100.0 & 61.5 & 0.0 & 0.0 & 0.0 & 27.7 & .001 & -- & -- \\
Llama 3.2-11B & T1a & 70.3 & 33.4 & 0.0 & 0.0 & 0.0 & 16.4 & .228 & -- & -- \\
Qwen3-VL-8B & T1a & 100.0 & 0.1 & 0.0 & 0.0 & 0.0 & 55.8 & .047 & -- & -- \\
\midrule
AnalogSeeker & T1b & 87.5 & 40.9 & 0.3 & 34.7 & 0.0 & 12.8 & .010 & 25.4 & 14.7 \\
Claude 4.5 & T1b & 100.0 & 77.8 & 0.0 & 39.2 & 0.0 & 15.0 & .000 & \textbf{47.8} & \textbf{38.5} \\
GPT-5.4-mini & T1b & 100.0 & 50.6 & 0.0 & 0.0 & 0.0 & 18.8 & .000 & 0.0 & 0.0 \\
Llama 3.2-11B & T1b & 94.9 & 10.2 & 0.4 & 0.9 & 0.0 & 8.7 & .000 & 2.2 & 0.03 \\
Qwen3-VL-8B & T1b & 98.8 & 36.6 & 0.0 & 12.9 & 0.0 & 20.2 & .074 & 7.5 & 0.5 \\
\bottomrule
\end{tabular}
\caption{\textbf{Full-10K T1a/T1b generation metrics}. This expands the compact main-paper view with the complete generation diagnostic family; rates and overlap are percentages.}
\label{tab:t1-full10k}
\end{table*}

\begin{table*}[t]
\centering
\small
\setlength{\tabcolsep}{3.8pt}
\begin{tabular}{@{}llrrrrrrrr@{}}
\toprule
Model & Input & $n$ & Prec. $\uparrow$ & Rec. $\uparrow$ & F1 $\uparrow$ & IoU $\uparrow$ & Hamming $\downarrow$ & Exact $\uparrow$ & Area diff. $\downarrow$ \\
\midrule
AnalogSeeker & primitive & 10000 & 28.7 & 9.7 & 14.6 & 7.9 & 22.6 & 0.0 & 16.5 \\
Claude 4.5 & primitive & 10000 & 26.5 & 38.0 & 31.2 & 18.5 & 33.1 & 0.0 & 11.8 \\
 & image & 10000 & 30.0 & 55.7 & 39.0 & 24.2 & 34.4 & 0.0 & 17.1 \\
GPT-5.4-mini & primitive & 10000 & 36.7 & 25.4 & 30.0 & 17.7 & 23.9 & 0.0 & 14.4 \\
 & image & 10000 & \textbf{48.2} & 47.2 & \textbf{47.7} & \textbf{31.3} & \textbf{20.5} & 0.0 & \textbf{7.5} \\
Llama 3.2-11B & primitive & 10000 & 19.8 & 21.0 & 20.4 & 11.3 & 43.5 & 0.0 & 32.0 \\
 & image & 10000 & 19.1 & 7.4 & 10.6 & 5.6 & 33.0 & 0.0 & 20.7 \\
Qwen3-VL-8B & primitive & 10000 & 28.7 & 0.6 & 1.2 & 0.6 & 19.9 & 0.0 & 19.8 \\
 & image & 10000 & 24.5 & 0.4 & 0.8 & 0.4 & 19.9 & 0.0 & 19.9 \\
\bottomrule
\end{tabular}
\caption{\textbf{Full-10K T2 representation-translation metrics} (percent), expanding the compact main-paper view with precision, recall, Hamming distance, exact match, and area error. AnalogSeeker has no archived T2b output.}
\label{tab:t2-full10k}
\end{table*}

\begin{table*}[t]
\centering
\small
\setlength{\tabcolsep}{3.8pt}
\begin{tabular}{@{}llrrrrrrrr@{}}
\toprule
Model & Input & Real MAE $\downarrow$ & Imag. MAE $\downarrow$ & Complex MAE $\downarrow$ & dB MAE $\downarrow$ & dB RMSE $\downarrow$ & $\leq$0.5 dB $\uparrow$ & $\leq$1 dB $\uparrow$ & $\leq$2 dB $\uparrow$ \\
\midrule
AnalogSeeker & primitive & .719 & .401 & .899 & 110.96 & 158.74 & 5.5 & 8.2 & 13.8 \\
Llama 3.2-11B & primitive & .405 & .402 & .636 & 99.81 & 138.55 & 0.1 & 0.1 & 0.3 \\
 & image & .447 & .401 & .673 & 80.09 & 126.44 & 0.8 & 1.6 & 3.1 \\
Qwen3-VL-8B & primitive & .642 & .401 & .831 & \textbf{21.29} & \textbf{26.89} & \textbf{6.8} & \textbf{11.4} & \textbf{17.6} \\
 & image & .526 & .402 & .732 & 32.05 & 50.94 & 3.9 & 7.1 & 11.3 \\
\bottomrule
\end{tabular}
\caption{\textbf{Available full-10K T3 response-prediction metrics}. Each displayed row has exactly 10,000 designs and uses all four $S$-parameter entries at 201 frequencies. Claude 4.5, GPT-5.4-mini, SFT, and GRPO have only official-panel T3 outputs and are not extrapolated.}
\label{tab:t3-full10k}
\end{table*}

\begin{table*}[t]
\centering
\small
\setlength{\tabcolsep}{5.0pt}
\begin{tabular}{@{}lrrrrrrr@{}}
\toprule
& & & \multicolumn{4}{c}{Specification-landmark MAE (dB) $\downarrow$} & \\
\cmidrule(lr){4-7}
Model & ADS runs & Run rate $\uparrow$ & All $S_{ij}$ & $S_{11}$ & $S_{12}=S_{21}$ & $S_{22}$ & Strict all $\leq3$ dB $\uparrow$ \\
\midrule
AnalogSeeker & 193/250 & 77.2 & 9.47 & 8.94 & 9.96 & 9.00 & 0.0 \\
Claude 4.5 & 219/250 & 87.6 & 7.26 & 8.34 & 6.19 & 8.34 & 0.4 \\
GPT-5.4-mini & 154/250 & 61.6 & \textbf{5.96} & 7.39 & \textbf{4.54} & 7.38 & 0.8 \\
Llama 3.2-11B & 93/250 & 37.2 & 10.55 & 8.19 & 12.92 & 8.18 & 0.4 \\
Qwen3-VL-8B & 217/250 & 86.8 & 9.74 & 8.31 & 11.16 & 8.33 & 0.0 \\
Qwen3 SFT & 250/250 & \textbf{100.0} & 9.65 & 7.02 & 12.26 & 7.05 & 1.6 \\
Qwen3 GRPO & 250/250 & \textbf{100.0} & 9.27 & \textbf{6.79} & 11.74 & \textbf{6.82} & \textbf{2.8} \\
\bottomrule
\end{tabular}
\caption{\textbf{Complete T4 closed-loop archive}. T4 is defined on the balanced 250-target subset, so no 10K T4 result exists. This reproduces the complete native-size table for task-by-task supplementary accounting.}
\label{tab:t4-fullarchive}
\end{table*}

\begin{table*}[t]
\centering
\small
\setlength{\tabcolsep}{4.0pt}
\begin{tabular}{@{}lrrrrrrrr@{}}
\toprule
& \multicolumn{5}{c}{T5: improvement diagnosis} & \multicolumn{3}{c}{T7: response classification} \\
\cmidrule(lr){2-6}\cmidrule(lr){7-9}
Model & $n$ & Acc. $\uparrow$ & Prec. $\uparrow$ & Rec. $\uparrow$ & Macro-F1 $\uparrow$ & $n$ & Acc. $\uparrow$ & Macro-F1 $\uparrow$ \\
\midrule
AnalogSeeker & 500 & 85.8 & 83.8 & 97.9 & 81.8 & 10000 & 8.5 & 7.9 \\
Claude 4.5 & 500 & 76.4 & 74.2 & \textbf{100.0} & 63.7 & 10000 & \textbf{48.7} & \textbf{37.1} \\
GPT-5.4-mini & 500 & \textbf{88.8} & \textbf{95.5} & 87.6 & \textbf{87.7} & 10000 & 24.6 & 16.1 \\
Llama 3.2-11B & 500 & 67.4 & 67.7 & 99.4 & 40.3 & 10000 & 27.8 & 7.2 \\
Qwen3-VL-8B & 500 & 52.4 & 65.3 & 63.4 & 46.3 & 10000 & 17.0 & 14.9 \\
Qwen3 SFT & 500 & 65.8 & 73.3 & 77.9 & 59.4 & -- & -- & -- \\
Qwen3 GRPO & 100 & 63.0 & 67.5 & 81.2 & 55.5 & -- & -- & -- \\
\bottomrule
\end{tabular}
\caption{\textbf{Combined full-archive T5/T7 classification metrics} (percent). T5 contains 500 cases rather than 10,000; GRPO has only its official 100-case output. The T7 columns report complete 10K outputs; SFT and GRPO have only official-panel T7 outputs and are not extrapolated.}
\label{tab:t5-t7-fullarchive}
\end{table*}

\begin{table*}[t]
\centering
\small
\setlength{\tabcolsep}{3.5pt}
\begin{tabular}{@{}llrrrrrrr@{}}
\toprule
Model & Input & $n$ & Cat. acc. $\uparrow$ & Cat. F1 $\uparrow$ & Mean feat. $\uparrow$ & Exact feat. $\uparrow$ & Joint all $\uparrow$ & Diag. $\uparrow$ \\
\midrule
AnalogSeeker & primitive & 10000 & 20.7 & 9.9 & 55.4 & 7.0 & 7.0 & 100.0 \\
Claude 4.5 & primitive & 10000 & \textbf{51.4} & \textbf{48.0} & 61.8 & 17.7 & 17.7 & 100.0 \\
 & image & 10000 & 47.6 & 46.6 & 67.0 & 13.9 & 13.0 & 100.0 \\
GPT-5.4-mini & primitive & 10000 & 22.8 & 13.6 & 55.7 & 15.7 & 15.7 & 100.0 \\
 & image & 10000 & 25.5 & 15.7 & 58.4 & 19.1 & 19.1 & 100.0 \\
Llama 3.2-11B & primitive & 10000 & 19.0 & 12.0 & \textbf{72.6} & 16.4 & 16.2 & 100.0 \\
 & image & 10000 & 20.1 & 7.5 & 56.5 & 13.2 & 13.2 & 97.2 \\
Qwen3-VL-8B & primitive & 10000 & 22.6 & 18.5 & 60.9 & 6.8 & 6.8 & 88.7 \\
 & image & 10000 & 35.5 & 23.8 & 69.9 & \textbf{35.5} & \textbf{35.5} & 100.0 \\
\bottomrule
\end{tabular}
\caption{\textbf{Full-10K T6 structural-understanding metrics} (percent), using the same feature definitions as the main-paper official-panel table. AnalogSeeker has no archived T6b output.}
\label{tab:t6-full10k}
\end{table*}

\FloatBarrier
\begin{table*}[p]
\begin{minipage}[t]{0.485\textwidth}
\section{Split Accounting and Adaptation Details}

The seed-42 design split holds out 1,000 designs, exactly 200 from each structural family. The remaining 9,000 designs are available for supervised training. T5 has its own fixed 100-case evaluation split, containing 64 positive and 36 negative \emph{needs-improvement} labels. No reported adaptation method trains on either official evaluation split. A separate seed-66 sampler selects the 1,000 GRPO design IDs from the non-test pool.

\paragraph{SFT objective.}
For prompt $x_i$ and target $y_i$, Qwen3 SFT masks prompt tokens and minimizes
$\mathcal L_{\mathrm{SFT}}=-N_y^{-1}\sum_{i,t}\log p_\theta(y_{i,t}\mid x_i,y_{i,<t})$, where $N_y=\sum_i|y_i|$.

\paragraph{Executed training commands.}
The archived wrappers below are the exact entry points used for the reported checkpoints. They activate the recorded environments and expand to the model, data, optimizer, LoRA, length, rollout, and checkpoint arguments in Table~\ref{tab:adaptation}; the expanded scripts, immutable VERL commit, package freezes, and checkpoint metadata are stored beside the released adapters.

\begin{small}
\begin{verbatim}
# Qwen3 SFT
cd RFbench_sft
bash outputs/qwen3vl8b_lora_exp1_eval2k_\
configs_scripts/run_sft_qwen3.sh
cd ..

# Qwen3 GRPO
cd RFbench_GRPO_backup_step_1100
bash code_and_configs/run_grpo_full_b2_n4.sh
\end{verbatim}
\end{small}

\end{minipage}
\hfill
\begin{minipage}[t]{0.485\textwidth}
The SFT wrapper invokes \texttt{train\_lora\_sft.py} for 14,063 optimizer updates (one epoch, effective batch 16). The GRPO wrapper invokes the pinned VERL 0.8.0 trainer at commit \texttt{7aed6b2} for 1,100 updates on two GPUs with four rollouts per prompt. The RF surrogate is trained on 8,100 non-test designs and selected on a disjoint 900-design validation set. Its selected checkpoint has validation real/imaginary component MAE 0.119 and magnitude MAE 1.449~dB. MLLM benchmark outputs and T4 EM results are computed independently of that surrogate; its separate official-panel T3 reference is reported in Table~\ref{tab:t3-official-detail}.

\paragraph{Computing infrastructure.}
The adaptation experiments ran in Linux containers on RunPod. SFT used one NVIDIA H100 SXM GPU, and GRPO used two such GPUs; each GPU provided 80~GB of VRAM (79.25~GiB usable), with 125~GB of host RAM and 8 vCPUs provisioned per GPU allocation. RunPod did not expose the physical host-CPU model. The recorded environment used Python~3.12.3, CUDA~12.8 runtime packages, PyTorch~2.9.0, Transformers~4.57.6, PEFT~0.19.1, and VERL~0.8.0 at commit \texttt{7aed6b2}. Dataset simulation used ADS Momentum on a 72-core Windows server.

\section{Paired Evaluation-Set Uncertainty}

Because SFT and GRPO are evaluated on matched examples, we estimate evaluation-set uncertainty with 5,000 paired bootstrap resamples using seed 2027. The resampling unit is one design for T1/T2/T3/T6/T7, one case for T5, and one target design for T4. Each sampled unit remains intact: in particular, all 201 frequencies and four $S_{ij}$ entries for a design are kept together rather than resampled independently. The same sampled IDs are used for SFT and GRPO within every replicate. Table~\ref{tab:adaptation-bootstrap} reports percentile intervals for $\Delta=\text{GRPO}-\text{SFT}$.

Every displayed interval includes or touches zero. These intervals quantify paired evaluation-panel sampling uncertainty for the fixed checkpoints, not variation across independent training runs.

\end{minipage}
\end{table*}

\begin{table*}[t]
\centering
\small
\setlength{\tabcolsep}{5pt}
\begin{tabular}{@{}lll@{}}
\toprule
Setting & Qwen3 SFT & Qwen3 GRPO \\
\midrule
Initialization & Qwen3-VL-8B & Qwen3 SFT \\
Objective & Assistant-token CE & Group-relative reward \\
Train volume & 225,000 conversations & 11,000-row pool \\
Optimization & 1 epoch & 1,100 steps \\
Learning rate & $2\!\times\!10^{-4}$ & $1\!\times\!10^{-6}$ \\
Prompt batch & 1 (accumulation 16) & 2 ($G=4$ outputs) \\
LoRA rank / scaling & 16 / 32 & 16 / 32 \\
Visual tower & QLoRA model adaptation & Frozen \\
Maximum lengths & 16,384 total & 16,384 prompt; 4,096 response \\
Sampling & -- & $T=1.0$, top-$p=0.95$ \\
KL coefficient & -- & 0.01 \\
GRPO design IDs & -- & 1,000 non-test IDs \\
Per-label cap & -- & 1,000 rows \\
Excluded GRPO tasks & -- & T1b and T4 \\
T5 train cases & None & 400 unique; 1,000 effective \\
\bottomrule
\end{tabular}
\caption{Adaptation settings. Training volumes differ by method, but all reported scores use identical official test IDs.}
\label{tab:adaptation}
\end{table*}

\begin{table*}[t]
\centering
\small
\setlength{\tabcolsep}{4.5pt}
\begin{tabular}{@{}llrrrl@{}}
\toprule
Task & Metric & SFT & GRPO & $\Delta$ & Paired 95\% CI for $\Delta$ \\
\midrule
T1a & Structural validity & 82.70 & 83.20 & $+0.50$ & $[-1.10,\ 2.10]$ \\
T1b & Structural validity & 86.00 & 86.40 & $+0.40$ & $[-1.50,\ 2.30]$ \\
T2a & Foreground F1 & 99.991 & 99.993 & $+0.002$ & $[0.000,\ 0.006]$ \\
T2b & Foreground F1 & 99.976 & 99.974 & $-0.002$ & $[-0.006,\ 0.000]$ \\
T3a & Within-2-dB point rate & 62.56 & 62.80 & $+0.24$ & $[-0.05,\ 0.53]$ \\
T3b & Within-2-dB point rate & 64.10 & 64.12 & $+0.01$ & $[-0.30,\ 0.32]$ \\
T5 & Decision macro-F1 & 58.64 & 55.52 & $-3.12$ & $[-10.73,\ 4.01]$ \\
T6a & Category macro-F1 & 99.40 & 99.30 & $-0.10$ & $[-0.32,\ 0.00]$ \\
T6b & Category macro-F1 & 98.80 & 99.00 & $+0.20$ & $[0.00,\ 0.51]$ \\
T7 & Response macro-F1 & 74.01 & 74.19 & $+0.19$ & $[0.00,\ 0.62]$ \\
T4 & Strict all-$S_{ij}$ success & 1.60 & 2.80 & $+1.20$ & $[0.00,\ 2.80]$ \\
\bottomrule
\end{tabular}
\caption{Paired evaluation-set bootstrap for the adapted checkpoints using 5,000 resamples and seed 2027. SFT, GRPO, differences, and percentile intervals are in percentage points. One design or case is the resampling unit, and all within-design frequency points remain grouped.}
\label{tab:adaptation-bootstrap}
\end{table*}

\clearpage
\begin{table*}[p]
\begin{minipage}[t]{0.485\textwidth}
\section{Exploratory RQS Profile Definition and Sensitivity}
We investigate RQS as an exploratory, coverage-aware capability-profile summary rather
than a primary benchmark metric. Native task metrics remain primary, and RQS is not used
to rank SFT against GRPO. Each native configuration is mapped to $q_t\in[0,1]$: T1a uses
validity; T1b uses the geometric mean of validity and full measurable compliance; T2 uses
foreground F1; T3 uses the within-2-dB rate; and T5--T7 use macro-F1. Primitive/image
variants are averaged within ability families to obtain $Q_f$, preventing paired modalities
from changing the weight of an ability. With observed families $\mathcal O$ and
completed-configuration fraction $C$, the reported profile is
\begin{equation}
\mathrm{RQS}=100C\left[1-\sqrt{\frac{1}{|\mathcal O|}
\sum_{f\in\mathcal O}(1-Q_f)^2}\right].
\label{eq:rqs-profile}
\end{equation}
\end{minipage}
\hfill
\begin{minipage}[t]{0.485\textwidth}
The root-mean-square deficiency gives greater influence to weak capabilities, equal family
weighting prevents paired modalities from being counted twice, and $C$ penalizes incomplete
evaluation. These are explicit design choices rather than a uniquely privileged scientific
scale. Table~\ref{tab:rqs-sensitivity} therefore varies the aggregator, treats all ten
configurations equally, leaves out each family in turn, changes the influence of the small
T5 panel, and incorporates T4 strict specification success. All alternatives use the same
normalized configuration metrics; no alternative is selected from performance.

The leave-one-family-out analysis retains the SFT point advantage when design,
translation, prediction, understanding, or classification is omitted; removing the 100-case
diagnosis family yields a small $+0.12$-point GRPO difference. The ordering is therefore
sensitive to the diagnosis contribution, while none of the paired intervals establishes a
nonzero aggregate difference. We consequently interpret RQS as a descriptive profile, not a
calibrated scientific distance or a substitute for the native task metrics.
\end{minipage}

\vspace{1.2em}
\centering
\small
\setlength{\tabcolsep}{4.0pt}
\begin{tabular}{@{}lrrrl@{}}
\toprule
Aggregation & SFT & GRPO & $\Delta$ & Paired 95\% CI for $\Delta$ \\
\midrule
Reported RMS-deficiency RQS & 74.25 & 73.48 & $-0.77$ & $[-2.88,\ 1.18]$ \\
Arithmetic mean of six families & 79.89 & 79.47 & $-0.42$ & $[-1.71,\ 0.78]$ \\
Geometric mean of six families & 78.25 & 77.64 & $-0.61$ & $[-2.42,\ 1.02]$ \\
Arithmetic mean of ten configurations & 82.60 & 82.39 & $-0.21$ & $[-1.02,\ 0.54]$ \\
Leave out design & 72.68 & 71.79 & $-0.90$ & $[-3.26,\ 1.31]$ \\
Leave out translation & 71.79 & 70.95 & $-0.84$ & $[-3.15,\ 1.29]$ \\
Leave out prediction & 77.05 & 75.98 & $-1.07$ & $[-3.86,\ 1.53]$ \\
Leave out diagnosis (T5) & 78.70 & 78.82 & $+0.12$ & $[-0.09,\ 0.34]$ \\
Leave out understanding & 71.79 & 70.95 & $-0.84$ & $[-3.15,\ 1.29]$ \\
Leave out classification & 74.30 & 73.34 & $-0.96$ & $[-3.49,\ 1.37]$ \\
RQS with half / double T5 weight & 76.17 / 71.49 & 75.76 / 70.24 & $-0.41$ / $-1.25$ & $[-1.70,\ 0.75]$ / $[-4.46,\ 1.73]$ \\
RQS plus T4 strict success & 55.82 & 55.81 & $-0.01$ & $[-1.17,\ 1.07]$ \\
\bottomrule
\end{tabular}
\caption{Sensitivity of the adapted-model aggregate comparison. $\Delta=\text{GRPO}-\text{SFT}$ in score points. Among the six leave-one-family-out variants, only omitting diagnosis changes the point ordering, and every paired interval spans zero. RQS therefore describes capability profiles and coverage but does not resolve the SFT--GRPO ordering. Exact results for 14 schemes are in \texttt{evaluation/reported\_results/rqs\_sensitivity.csv}.}
\label{tab:rqs-sensitivity}
\end{table*}

\clearpage
\begin{table*}[p]
\begin{minipage}[t]{0.485\textwidth}
\section{RQS Profile Decomposition and Coverage}
Table~\ref{tab:rqs-decomposition} expands Equation~\ref{eq:rqs-profile} into its six ability-family inputs. Each value is the mapped score $Q_f$, expressed as a percentage. Primitive and image configurations are averaged within translation, prediction, and understanding. AnalogSeeker has no archived image prediction for T2b, T3b, or T6b; its displayed family value uses the observed primitive configuration, while its $7/10$ coverage multiplies the final score by 0.7. T4 is reported in its own closed-loop table; Table~\ref{tab:rqs-sensitivity} shows the effect of including it.

The decomposition makes the profile interpretable. Qwen3 GRPO has slightly higher point estimates in design, prediction, understanding, and classification, while Qwen3 SFT retains a 3.1-point diagnosis advantage. Under the reported RMS-deficiency form, that diagnosis difference changes the point score from 74.2 to 73.5; the paired interval and alternative aggregations do not resolve an adapted-model ordering. Among unadapted models, Claude 4.5's broader translation, prediction, and understanding performance produces the highest RQS even though GPT-5.4-mini is substantially stronger on diagnosis.
\end{minipage}
\hfill
\begin{minipage}[t]{0.485\textwidth}
\section{GRPO Reward and Evaluation Alignment}
The policy objective uses per-sample training rewards, whereas the official benchmark uses held-out aggregate metrics. Table~\ref{tab:reward-alignment} records the exact alignment. A reward can therefore provide useful local learning signal without being identical to the configuration score used by RQS.

This separation is essential when interpreting GRPO. The pool has 11,000 rows drawn from 1,000 non-test designs, and its T5 portion repeats 400 unique cases to reach 1,000 effective rows. T5 reward values both the decision and a proposed redesign, whereas the official metric evaluates only decision macro-F1. The frozen surrogate supplies a smooth training signal for T1a/T5, but no MLLM benchmark score uses its predictions as ground truth or as a scoring function. Consequently, the reported evidence is the held-out task profile and EM execution result, not the training reward itself; Table~\ref{tab:adaptation-bootstrap} gives the corresponding paired uncertainty.
\end{minipage}

\vspace{1.2em}
\centering
\small
\setlength{\tabcolsep}{4.0pt}
\begin{tabular}{@{}lcrrrrrrr@{}}
\toprule
Model & Coverage & Design & Translation & Prediction & Diagnosis & Understanding & Classification & RQS \\
\midrule
AnalogSeeker & 7/10 & 0.0 & 14.2 & 13.4 & 82.0 & 10.9 & 7.5 & 11.7 \\
Claude 4.5 & 10/10 & 0.0 & 35.4 & 27.5 & 61.3 & 46.7 & 36.4 & 31.9 \\
GPT-5.4-mini & 10/10 & 0.0 & 38.5 & 20.0 & \textbf{88.4} & 14.5 & 16.8 & 24.1 \\
Llama 3.2-11B & 10/10 & 0.0 & 15.4 & 1.7 & 39.0 & 9.9 & 7.3 & 11.2 \\
Qwen3-VL-8B & 10/10 & 0.0 & 0.8 & 14.4 & 51.0 & 22.1 & 14.8 & 15.5 \\
Qwen3 SFT & 10/10 & 84.3 & \textbf{100.0} & 63.3 & 58.6 & 99.1 & 74.0 & \textbf{74.2} \\
Qwen3 GRPO & 10/10 & \textbf{84.5} & \textbf{100.0} & \textbf{63.5} & 55.5 & \textbf{99.1} & \textbf{74.2} & 73.5 \\
\bottomrule
\end{tabular}
\caption{RQS decomposition on the fixed official evaluation IDs. Family scores and RQS are percentages; displayed values are rounded to one decimal. Values are descriptive point estimates; paired uncertainty and alternative aggregations are reported in Table~\ref{tab:rqs-sensitivity}. The exact inputs are released in \texttt{evaluation/reported\_results/rqs\_summary.csv}.}
\label{tab:rqs-decomposition}

\vspace{1.0em}
\small
\setlength{\tabcolsep}{4.2pt}
\begin{tabular}{@{}p{1.05cm}p{9.15cm}@{\hspace{0.25cm}}p{5.3cm}@{}}
\toprule
Task & GRPO training reward & Official RQS input \\
\midrule
T1a & Invalid: $0.15v$. Valid: $0.25v+0.75z$, with graded geometry $v$ and surrogate specification score $z$. & Structural validity. \\
T2 & $0.75$ foreground F1 $+0.25$ exact-token match. & Foreground F1, averaged over T2a/T2b. \\
T3 & Auxiliary response-type rows use exact class correctness; real/imaginary rows use clipped $1-\mathrm{MAE}/0.25$. & Fraction of complex $S$-parameter points within 2~dB, averaged over T3a/T3b. \\
T5 & Piecewise decision reward; a valid necessary redesign uses $0.20v+0.55z+0.25\operatorname{clip}(0.5+z-z_0,0,1)$. & Macro-F1 of the binary \emph{needs-improvement} decision. \\
T6 & $0.70$ category correctness $+0.30$ mean feature correctness. & Category macro-F1, averaged over T6a/T6b. \\
T7 & Exact response-type correctness. & Six-class response macro-F1. \\
T1b/T4 & Not included in the GRPO pool. & T1b uses mapped validity/compliance; T4 reports EM run and specification success separately from RQS. \\
\bottomrule
\end{tabular}
\caption{Relationship between sample-level GRPO rewards and held-out RQS inputs. The frozen RF surrogate shapes T1a/T5 rewards but is not used to generate or score official MLLM predictions; its separate grid-input T3 reference is reported in Table~\ref{tab:t3-official-detail}.}
\label{tab:reward-alignment}
\end{table*}

\clearpage
\section{Exact Inference Prompts and Output Schemas}

All benchmark model inputs are issued through natural-language instructions. For text-input
tasks, the serialized primitive, specification, or $S$-parameter payload replaces the
brace-delimited placeholder in the user template. For image-input tasks, the same instruction
is paired with the referenced layout image. The blocks below reproduce the literal prompt
constants used by the released synchronous inference scripts; braces denote run-time
substitutions. T4 has no model prompt because it imports T1a outputs into ADS Momentum and
scores the resulting simulation.
Thus the model-called tasks documented below are T1a, T1b, T2a, T2b, T3a,
T3b, T5, T6a, T6b, and T7. The T4 non-prompt protocol and evaluator record
are documented in task order below so that every row of the main task taxonomy
has an explicit input/output account.

\subsection{T1a: Specification-Conditioned Generation}

\paragraph{System prompt.}
\begin{Verbatim}[fontsize=\scriptsize,breaklines=true,breakanywhere=true]
You generate RF/microstrip-inspired layout primitives for benchmark inference.
Return only primitive lines. Do not include SVG wrappers, Markdown fences, explanations, comments, or extra text.
\end{Verbatim}

\paragraph{User template and output schema.}
\begin{Verbatim}[fontsize=\scriptsize,breaklines=true,breakanywhere=true]
Task: Generate one RF/microstrip-inspired layout from the target RF/S-parameter specification below.

The specification is a structured RF/S-parameter analysis JSON. It may include
response type, key transmission extrema, passband summaries, frequency windows,
sparse S-parameter sample points, and grouped point sets. Use it as the target
response signature for the layout.

Output format:
- Return primitive lines only, one element per line.
- Do not return a complete <svg> wrapper.
- Use rect primitives for transmission-line or metal regions:
  rect x="..." y="..." width="..." height="..."
- Use circle primitives for port markers:
  circle id="port1" cx="..." cy="..." r="..."
  circle id="port2" cx="..." cy="..." r="..."
- You may include one canvas/background rect as the first line.
- Do not include fill, stroke, class, style, or transform attributes.
- Choose the canvas, port positions, and trace geometry according to the target RF/S-parameter specification.
- Keep the layout concise, but avoid returning the same generic through-line for different RF/S-parameter specifications; meaningfully different RF signatures should generally produce different primitive arrangements.
- Prefer simple axis-aligned rect geometry and grid-like coordinates when appropriate.

Target RF/S-parameter specification:
{spec}
\end{Verbatim}

\subsection{T1b: Category-Constrained Generation}

\paragraph{System prompt.}
\begin{Verbatim}[fontsize=\scriptsize,breaklines=true,breakanywhere=true]
You generate RF/microstrip-inspired layout primitives for benchmark inference.
Return only primitive lines. Do not include SVG wrappers, Markdown fences, explanations, comments, or extra text.
\end{Verbatim}

\paragraph{User template and output schema.}
\begin{Verbatim}[fontsize=\scriptsize,breaklines=true,breakanywhere=true]
Task: Generate one RF/microstrip-inspired layout from the target RF/S-parameter specification below.

The specification is a structured RF/S-parameter analysis JSON. It may include
response type, key transmission extrema, passband summaries, frequency windows,
sparse S-parameter sample points, and grouped point sets. Use it as the target
response signature for the layout.

Output format:
- Return primitive lines only, one element per line.
- Do not return a complete <svg> wrapper.
- Use rect primitives for transmission-line or metal regions:
  rect x="..." y="..." width="..." height="..."
- Use circle primitives for port markers:
  circle id="port1" cx="..." cy="..." r="..."
  circle id="port2" cx="..." cy="..." r="..."
- You may include one canvas/background rect as the first line.
- Do not include fill, stroke, class, style, or transform attributes.
- Choose the canvas, port positions, and trace geometry according to the target RF/S-parameter specification.
- Keep the layout concise, but avoid returning the same generic through-line for different RF/S-parameter specifications; meaningfully different RF signatures should generally produce different primitive arrangements.
- Prefer simple axis-aligned rect geometry and grid-like coordinates when appropriate.

Required category:
{category}

Category requirement:
{category_requirement}

Target RF/S-parameter specification:
{spec}
\end{Verbatim}

The \texttt{category\_requirement} placeholder is replaced by the exact matching entry below.
\begin{Verbatim}[fontsize=\scriptsize,breaklines=true,breakanywhere=true]
edge_connected_random:
The layout is an edge-connected random design: metal regions form a feed-to-feed connected random-walk-like structure with random fill or noise-like texture.

microstrip_primitives:
The layout is built from microstrip-inspired primitives: it has a clear RF-style center trace and may include intentional stub or resonator-like features.

rot180:
The layout must have 180-degree rotational symmetry about (0, 0). For every rect at (x, y) with width w and height h, there must be a corresponding rect at (-x - w, -y - h) with the same width and height.

symmetric_lr:
The layout must be left-right (mirror) symmetric about x=0. For every rect at (x, y) with width w and height h, there must be a corresponding rect at (-x - w, y) with the same width and height.

symmetric_tb:
The layout must be top-bottom (mirror) symmetric about y=0. For every rect at (x, y) with width w and height h, there must be a corresponding rect at (x, -y - h) with the same width and height.
\end{Verbatim}

\subsection{T2a: Primitive-to-Grid Translation}

\paragraph{System prompt.}
\begin{Verbatim}[fontsize=\scriptsize,breaklines=true,breakanywhere=true]
You convert RF/microstrip layout primitive text into binary geometry tokens.
Return only the binary token grid. Do not include explanations, Markdown fences, spaces, row numbers, or labels.
\end{Verbatim}

\paragraph{User template and output schema.}
\begin{Verbatim}[fontsize=\scriptsize,breaklines=true,breakanywhere=true]
Task: Convert the RF/microstrip rect/circle primitive layout into a binary geometry token.

Output requirements:
- Return exactly {rows} rows.
- Each row must contain exactly {cols} characters.
- Use only 0 and 1.
- 1 means the token cell is occupied by non-background layout geometry.
- 0 means empty space or background.
- Ignore the background/canvas.
- Do not mark port markers as occupied cells.

Primitive input interpretation:
- Rect primitives define the layout geometry.
- Circle primitives are port markers only.
- Ignore any first large rect that represents only the background canvas.
- The default 16x16 grid uses 1.25-unit cells.
- Columns go left to right from x=-10, -8.75, ..., 8.75.
- Rows go top to bottom from y=8.75, 7.5, ..., -10.
- A rect marks every grid cell covered by its x/y/width/height span.
- Port-feed rects outside the grid that touch x=-10 or x=10 should mark the left or right edge cell at that row.

Primitive geometry input:
{primitives}
\end{Verbatim}

\subsection{T2b: Image-to-Grid Translation}

\paragraph{System prompt.}
\begin{Verbatim}[fontsize=\scriptsize,breaklines=true,breakanywhere=true]
You convert RF/microstrip layout images into binary geometry tokens.
Return only the binary token grid. Do not include explanations, Markdown fences, spaces, row numbers, or labels.
\end{Verbatim}

\paragraph{User template and output schema.}
\begin{Verbatim}[fontsize=\scriptsize,breaklines=true,breakanywhere=true]
Task: Convert the RF/microstrip layout image into a binary geometry token.

Output requirements:
- Return exactly {rows} rows.
- Each row must contain exactly {cols} characters.
- Use only 0 and 1.
- 1 means the token cell is occupied by non-background layout geometry.
- 0 means empty space or background.
- Ignore the background/canvas.
- Do not mark port markers as occupied cells.

Image input interpretation:
- Use the visible non-background layout geometry in the image.
- Ignore the white/background canvas.
- Do not mark circle port markers as occupied cells.
\end{Verbatim}

\subsection{T3a: Primitive-to-Response Prediction}

\paragraph{System prompt.}
\begin{Verbatim}[fontsize=\scriptsize,breaklines=true,breakanywhere=true]
You are an RF surrogate model for two-port microstrip-like layouts.
Estimate raw S-parameter sweeps from primitive SVG-style geometry and requested frequency points.
Return only JSON. Do not include Markdown fences, comments, or text outside JSON.
Do not run or claim to run EM simulation.
\end{Verbatim}

\paragraph{Response-type user template and schema.}
\begin{Verbatim}[fontsize=\scriptsize,breaklines=true,breakanywhere=true]
Task: Classify the overall RF response type for this primitive design.

Full requested frequency span in GHz:
{frequency_ghz}

Input:
Primitive geometry input:
{primitives}

{response_type_definitions}

Output requirements:
- Return one JSON object only.
- Classify the overall S21 response across the full requested frequency span.
- Use exactly one response_type label.

JSON schema:
{{
  "response_type": "broadband_through" | "broadband_attenuating" | "lowpass" | "bandpass" | "bandstop_notch" | "resonant_or_shaped"
}}
\end{Verbatim}

\paragraph{Real/imaginary segment user template and schema.}
\begin{Verbatim}[fontsize=\scriptsize,breaklines=true,breakanywhere=true]
Task: Predict one frequency segment of the real and imaginary S-parameter sweep for this primitive design.

Full requested frequency span in GHz:
{full_frequency_ghz}

Frequency segment to predict in GHz:
{frequency_ghz}

This segment contains exactly {frequency_count} frequency points.

Input:
Primitive geometry input:
{primitives}

{response_type_definitions}

Global response context:
The overall response_type is {response_type}. Predict this frequency segment consistently with that global response.

Output requirements:
- Return one JSON object only.
- Use exactly the frequency segment list and preserve its order.
- Every S-parameter array must have exactly {frequency_count} numeric values.
- Predict all four two-port S-parameters: S11, S12, S21, and S22.
- Do not include response_type in this segment JSON. It is provided separately as global context.
- s_real and s_imag are the real and imaginary components of the complex S-parameter.
- Use numeric values only. Do not use null.
- Round every s_real and s_imag value to 2 decimal places.
- Do not include values for frequencies outside this segment.
- Stop immediately after the closing JSON object.

JSON schema:
{{
  "frequency_ghz": [number, ...],
  "s_real": {{
    "S11": [number, ...],
    "S12": [number, ...],
    "S21": [number, ...],
    "S22": [number, ...]
  }},
  "s_imag": {{
    "S11": [number, ...],
    "S12": [number, ...],
    "S21": [number, ...],
    "S22": [number, ...]
  }}
}}
\end{Verbatim}

\subsection{T3b: Image-to-Response Prediction}

\paragraph{System prompt.}
\begin{Verbatim}[fontsize=\scriptsize,breaklines=true,breakanywhere=true]
You are an RF surrogate model for two-port microstrip-like layouts.
Estimate raw S-parameter sweeps from the provided layout image and requested frequency points.
Return only JSON. Do not include Markdown fences, comments, or text outside JSON.
Do not run or claim to run EM simulation.
\end{Verbatim}

\paragraph{Response-type user template and schema.}
\begin{Verbatim}[fontsize=\scriptsize,breaklines=true,breakanywhere=true]
Task: Classify the overall RF response type for this layout image.

Full requested frequency span in GHz:
{frequency_ghz}

Input:
The layout image is provided as the image input for this request.

{response_type_definitions}

Output requirements:
- Return one JSON object only.
- Classify the overall S21 response across the full requested frequency span.
- Use exactly one response_type label.

JSON schema:
{{
  "response_type": "broadband_through" | "broadband_attenuating" | "lowpass" | "bandpass" | "bandstop_notch" | "resonant_or_shaped"
}}
\end{Verbatim}

\paragraph{Real/imaginary segment user template and schema.}
\begin{Verbatim}[fontsize=\scriptsize,breaklines=true,breakanywhere=true]
Task: Predict one frequency segment of the real and imaginary S-parameter sweep for this layout image.

Full requested frequency span in GHz:
{full_frequency_ghz}

Frequency segment to predict in GHz:
{frequency_ghz}

This segment contains exactly {frequency_count} frequency points.

Input:
The layout image is provided as the image input for this request.

{response_type_definitions}

Global response context:
The overall response_type is {response_type}. Predict this frequency segment consistently with that global response.

Output requirements:
- Return one JSON object only.
- Use exactly the frequency segment list and preserve its order.
- Every S-parameter array must have exactly {frequency_count} numeric values.
- Predict all four two-port S-parameters: S11, S12, S21, and S22.
- Do not include response_type in this segment JSON. It is provided separately as global context.
- s_real and s_imag are the real and imaginary components of the complex S-parameter.
- Use numeric values only. Do not use null.
- Round every s_real and s_imag value to 2 decimal places.
- Do not include values for frequencies outside this segment.
- Stop immediately after the closing JSON object.

JSON schema:
{{
  "frequency_ghz": [number, ...],
  "s_real": {{
    "S11": [number, ...],
    "S12": [number, ...],
    "S21": [number, ...],
    "S22": [number, ...]
  }},
  "s_imag": {{
    "S11": [number, ...],
    "S12": [number, ...],
    "S21": [number, ...],
    "S22": [number, ...]
  }}
}}
\end{Verbatim}

For both T3 variants, the following literal block replaces
\texttt{\{response\_type\_definitions\}}. The response-type request is made once over the
full sweep; the real/imaginary request is then issued in 25-frequency-point segments
(the final segment absorbs the remainder), and the segment JSON objects are concatenated
in frequency order.
\begin{Verbatim}[fontsize=\scriptsize,breaklines=true,breakanywhere=true]
Allowed response_type labels:
- broadband_through: S21 is mostly low-loss and smooth across the requested band, without a strong filter edge or localized rejection feature.
- broadband_attenuating: S21 is weak/attenuated across most of the requested band, without one clear narrow passband or notch.
- lowpass: lower frequencies transmit better than higher frequencies, with a rolloff or rejection trend toward the high end of the band.
- bandpass: there is a localized passband or transmission peak, with weaker transmission on both lower and higher sides.
- bandstop_notch: there is a localized S21 rejection notch, while surrounding frequencies transmit better.
- resonant_or_shaped: S21 is non-flat or resonant, but does not cleanly match lowpass, bandpass, bandstop_notch, broadband_through, or broadband_attenuating.
\end{Verbatim}

\subsection{T4: ADS Execution Has No Model Prompt}

T4 does not make another language-model call and therefore has no system
prompt, user prompt, or model-output schema. It takes the frozen T1a primitive
output for each of the 250 preselected targets, imports that geometry into ADS
Momentum, and attempts the same 201-point solve used for the dataset. The
evaluator record joins the target \texttt{sample\_id}, \texttt{design\_id},
category, response label, selection index, and 12-value T1 specification to
the generated-layout identifier, model name, \texttt{simulation\_success}
flag, optional failure reason, and, when successful, the complete simulated
frequency and complex $S$-parameter arrays. Thus ``run/fail + simulated
response'' in the main task taxonomy describes an evaluator-produced record,
not text requested from a model. Equations~\ref{eq:t4-run-mae} and
\ref{eq:t4-strict} define how that record is scored.

\subsection{T5: Design Diagnosis and Improvement}

\paragraph{System prompt.}
\begin{Verbatim}[fontsize=\scriptsize,breaklines=true,breakanywhere=true]
You judge and improve RF/microstrip-inspired layout primitives using measured or simulated S-parameter results and a target spec.
Return only JSON. Do not include Markdown fences, comments, or text outside JSON.
\end{Verbatim}

\paragraph{User template and output schemas.}
The reported runs use \texttt{max\_reasoning\_words=50}.
\begin{Verbatim}[fontsize=\scriptsize,breaklines=true,breakanywhere=true]
Task: Decide whether the candidate layout needs improvement for the target RF/S-parameter specification. Only redesign it if improvement is needed.

Inputs:
- Current design: primitive geometry lines using rect elements for metal traces and circle elements for port markers.
- Current S-parameter result: measured, simulated, or predicted S-parameter text/JSON.
- Target RF/S-parameter specification: desired behavior.

Output requirements:
- Return one JSON object only.
- First decide whether the current design needs improvement.
- Use lowercase "yes" or "no" for needs_improvement.
- If needs_improvement is "no", set improved_design to [] and reasoning to "".
- If needs_improvement is "yes", provide an improved primitive design and a short reasoning string.
- Keep reasoning at {max_reasoning_words} words or fewer.
- Reasoning should explain why the proposed geometry helps satisfy the target RF/S-parameter spec.
- Keep the improved design in the same primitive format as the input.
- Keep the layout concise, but avoid returning the same generic through-line for different RF/S-parameter specifications; meaningfully different RF signatures should generally produce different primitive arrangements.
- Use rect primitives for transmission-line or metal regions:
  rect x="..." y="..." width="..." height="..."
- Use circle primitives for port markers:
  circle id="port1" cx="..." cy="..." r="..."
  circle id="port2" cx="..." cy="..." r="..."
- Do not include fill, stroke, class, style, or transform attributes.

JSON schema when no improvement is needed:
{{
  "needs_improvement": "no",
  "improved_design": [],
  "reasoning": ""
}}

JSON schema when improvement is needed:
{{
  "needs_improvement": "yes",
  "improved_design": [
    "rect x=\"...\" y=\"...\" width=\"...\" height=\"...\"",
    "circle id=\"port1\" cx=\"...\" cy=\"...\" r=\"...\""
  ],
  "reasoning": "brief explanation"
}}

Target RF/S-parameter specification:
{spec}

Current S-parameter result:
{sparameters}

Current design:
{design}
\end{Verbatim}

\subsection{T6a: Primitive Structural Understanding}

\paragraph{System prompt.}
\begin{Verbatim}[fontsize=\scriptsize,breaklines=true,breakanywhere=true]
You classify RF/microstrip-inspired primitive layouts into benchmark categories and detect structural features.
Return only JSON. Do not include Markdown fences, comments, or text outside JSON.
\end{Verbatim}

\paragraph{User template and output schema.}
\begin{Verbatim}[fontsize=\scriptsize,breaklines=true,breakanywhere=true]
Task: Classify the primitive layout and detect structural features.

Allowed labels:
- edge_connected_random
- microstrip_primitives
- rot180
- symmetric_lr
- symmetric_tb

Category requirements:
- edge_connected_random: {edge_connected_random}
- microstrip_primitives: {microstrip_primitives}
- rot180: {rot180}
- symmetric_lr: {symmetric_lr}
- symmetric_tb: {symmetric_tb}

Detected feature definitions:
- feed_to_feed_connected: there is a connected conductive path between port1 and port2.
- random_walk_backbone: the main trace turns and meanders instead of being mostly straight.
- rf_center_trace: the design has a clear RF-inspired central feed or center trace.
- stub_resonator_features: the design has intentional stub or resonator-like branches.
- left_right_symmetry: the rect layout has left-right mirror symmetry about x=0.
- top_bottom_symmetry: the rect layout has top-bottom mirror symmetry about y=0.
- rotation_180_symmetry: the rect layout has 180-degree rotational symmetry about (0, 0).
- random_fill_texture: the layout has random fill or noise-like metal texture.
- balanced_mirrored_structure: the layout appears balanced or mirrored by construction.
- diagonal_connected_blocks: metal blocks are connected only diagonally, through corners rather than shared edges.

Return one JSON object only, using this schema:
{{
  "category": "edge_connected_random" | "microstrip_primitives" | "rot180" | "symmetric_lr" | "symmetric_tb",
  "detected_features": {{
    "feed_to_feed_connected": true or false,
    "random_walk_backbone": true or false,
    "rf_center_trace": true or false,
    "stub_resonator_features": true or false,
    "left_right_symmetry": true or false,
    "top_bottom_symmetry": true or false,
    "rotation_180_symmetry": true or false,
    "random_fill_texture": true or false,
    "balanced_mirrored_structure": true or false,
    "diagonal_connected_blocks": true or false
  }}
}}

Primitive geometry input:
{primitives}
\end{Verbatim}

\subsection{T6b: Image Structural Understanding}

\paragraph{System prompt.}
\begin{Verbatim}[fontsize=\scriptsize,breaklines=true,breakanywhere=true]
You classify RF/microstrip-inspired layout images into benchmark categories and detect structural features.
Return only JSON. Do not include Markdown fences, comments, or text outside JSON.
\end{Verbatim}

\paragraph{User template and output schema.}
\begin{Verbatim}[fontsize=\scriptsize,breaklines=true,breakanywhere=true]
Task: Classify the layout image and detect structural features.

Allowed labels:
- edge_connected_random
- microstrip_primitives
- rot180
- symmetric_lr
- symmetric_tb

Category requirements:
- edge_connected_random: {edge_connected_random}
- microstrip_primitives: {microstrip_primitives}
- rot180: {rot180}
- symmetric_lr: {symmetric_lr}
- symmetric_tb: {symmetric_tb}

Detected feature definitions:
- feed_to_feed_connected: there is a connected conductive path between port1 and port2.
- random_walk_backbone: the main trace turns and meanders instead of being mostly straight.
- rf_center_trace: the design has a clear RF-inspired central feed or center trace.
- stub_resonator_features: the design has intentional stub or resonator-like branches.
- left_right_symmetry: the rect layout has left-right mirror symmetry about x=0.
- top_bottom_symmetry: the rect layout has top-bottom mirror symmetry about y=0.
- rotation_180_symmetry: the rect layout has 180-degree rotational symmetry about (0, 0).
- random_fill_texture: the layout has random fill or noise-like metal texture.
- balanced_mirrored_structure: the layout appears balanced or mirrored by construction.
- diagonal_connected_blocks: metal blocks are connected only diagonally, through corners rather than shared edges.

Return one JSON object only, using this schema:
{{
  "category": "edge_connected_random" | "microstrip_primitives" | "rot180" | "symmetric_lr" | "symmetric_tb",
  "detected_features": {{
    "feed_to_feed_connected": true or false,
    "random_walk_backbone": true or false,
    "rf_center_trace": true or false,
    "stub_resonator_features": true or false,
    "left_right_symmetry": true or false,
    "top_bottom_symmetry": true or false,
    "rotation_180_symmetry": true or false,
    "random_fill_texture": true or false,
    "balanced_mirrored_structure": true or false,
    "diagonal_connected_blocks": true or false
  }}
}}
\end{Verbatim}

For T6a and T6b, the five brace-delimited category placeholders are replaced by
the same literal category-requirement strings listed under T1b.

\subsection{T7: Response Classification}

\paragraph{System prompt.}
\begin{Verbatim}[fontsize=\scriptsize,breaklines=true,breakanywhere=true]
You classify two-port RF S-parameter responses into benchmark response types.
Return only JSON. Do not include Markdown fences, comments, or text outside JSON.
\end{Verbatim}

\paragraph{User template and output schema.}
\begin{Verbatim}[fontsize=\scriptsize,breaklines=true,breakanywhere=true]
Task: Classify the RF response type from the S-parameter data.

{response_type_definitions}

Input S-parameter data:
{sparameters}

Return one JSON object only, using this schema:
{{
  "response_type": "broadband_through" | "broadband_attenuating" | "lowpass" | "bandpass" | "bandstop_notch" | "resonant_or_shaped"
}}
\end{Verbatim}

The \texttt{\{response\_type\_definitions\}} placeholder is replaced by the same
literal six-label definition block reproduced under T3.

\end{document}